\pdfoutput=1

\documentclass[11pt]{article}

\usepackage[]{acl}

\usepackage{times}
\usepackage{latexsym}

\usepackage[T1]{fontenc}

\usepackage[utf8]{inputenc}

\usepackage{microtype}

\usepackage{inconsolata}

\usepackage{booktabs}
\usepackage{longtable}
\usepackage{tcolorbox}
\usepackage{stackengine}
\usepackage{multirow}
\usepackage{graphicx}
\usepackage{wrapfig}
\usepackage{hyperref}       
\usepackage{url}            
\usepackage{amsfonts}       
\usepackage{nicefrac}       
\usepackage{microtype}      
\usepackage{xcolor,colortbl}         
\usepackage{times}
\usepackage{latexsym}
\usepackage{multicol}
\usepackage{blindtext}
\usepackage{tabu}
\usepackage{amsmath, bm}

\usepackage{subcaption}
\usepackage{caption}
\usepackage[normalem]{ulem}
\usepackage{soul}
\usepackage{color}
\usepackage[shortlabels]{enumitem}
\usepackage{array}
\usepackage{pgffor}
\usepackage{textcomp}
\usepackage{amssymb}
\usepackage{pifont}
\newcommand{\xmark}{\ding{55}}
\definecolor{lightgreen}{rgb}{0.8, 0.95, 0.8}
\definecolor{lightred}{rgb}{0.95, 0.8, 0.8}
\definecolor{naplesyellow}{rgb}{0.98, 0.85, 0.37}
\definecolor{pastelyellow}{rgb}{0.99, 0.99, 0.59}
\DeclareRobustCommand{\hlred}[1]{{\sethlcolor{lightred}\hl{#1}}}

\title{LogicBench: Towards Systematic Evaluation of Logical Reasoning Ability of Large Language Models}

\author{Mihir Parmar$^1$ \quad Nisarg Patel$^1$ \quad Neeraj Varshney$^1$ \quad Mutsumi Nakamura$^1$ \quad \textbf{Man Luo}$^1$ \\ \quad \textbf{Santosh Mashetty}$^1$ \quad \textbf{Arindam Mitra}$^2$ \quad \textbf{Chitta Baral}$^1$ \\\\ 
$^1$Arizona State University \quad $^2$Microsoft Research\\
\small{\texttt{\{mparmar3, nppatel7, chitta\}@asu.edu}}
}

\begin{document}
\maketitle

\begin{abstract}

Recently developed large language models (LLMs) have been shown to perform remarkably well on a wide range of language understanding tasks. 
But, can they really ``{reason}'' over the natural language?
This question has been receiving significant research attention and many reasoning skills such as commonsense, numerical, and qualitative have been studied.
However, the crucial skill pertaining to `logical reasoning' has remained underexplored.
Existing work investigating this reasoning ability of LLMs has focused only on 
a couple of inference rules (such as modus ponens and modus tollens) of propositional and first-order logic. 
Addressing the above limitation, we comprehensively evaluate the logical reasoning ability of LLMs on 25 different reasoning patterns spanning over propositional, first-order, and non-monotonic logics.
To enable systematic evaluation, we introduce \textit{LogicBench}, a natural language question-answering dataset focusing on the use of a single inference rule.  
We conduct detailed analysis with a range of LLMs such as GPT-4, ChatGPT, Gemini, Llama-2, and Mistral using chain-of-thought prompting.
Experimental results show that existing LLMs do not fare well on \textit{LogicBench}; especially, they struggle with instances involving complex reasoning and negations. 
Furthermore, they sometimes overlook contextual information necessary for reasoning to arrive at the correct conclusion.
We believe that our work and findings facilitate future research for evaluating and enhancing the logical reasoning ability of LLMs\footnote{Data and code are available at \url{https://github.com/Mihir3009/LogicBench}}.

\end{abstract}

\section{Introduction}
\label{sec:intro}


\begin{figure}[!ht]
    \centering
    \includegraphics[width=\linewidth]{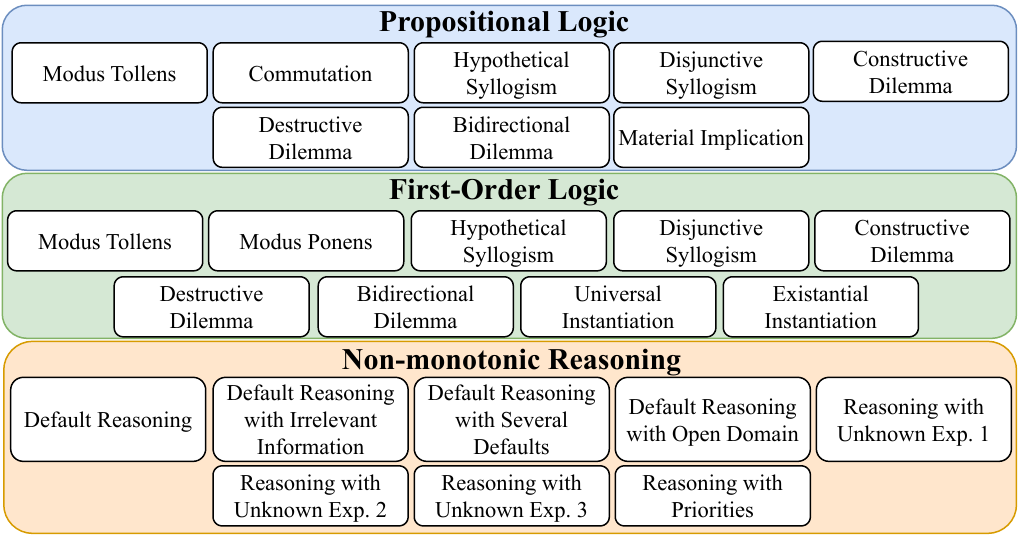}
    \caption{Comprehensive representation of different inference rules and reasoning patterns covered by propositional, first-order, and non-monotonic logics. \textit{Exp.} indicates Expectation}
    \label{fig:logic_type}
    \vspace{-0.2cm}
\end{figure}

Large language models such as GPT-4, ChatGPT, Google Gemini, Llama-2 \citep{touvron2023llama}, and Mistral \citep{jiang2023mistral} have made remarkable progress in NLP research enabling machines to perform a variety of language tasks that were previously thought to be exclusive to humans \citep{openai2023gpt4, NEURIPS2020_1457c0d6, zhao2023survey}. However, the ability of these LLMs to reason ``logically'' over natural language text remains under-explored, even though logical reasoning is a fundamental aspect of intelligence and a crucial requirement for many practical applications, such as question-answering systems \citep{khashabi2019reasoning} and conversational agents \citep{beygi-etal-2022-logical}.   
Although several datasets have been proposed \citep{clark2021transformers, tian-etal-2021-diagnosing, joshi-etal-2020-taxinli, saeed-etal-2021-rulebert} to evaluate the logical reasoning capabilities of LLMs, these datasets are limited in their scope by (1) not evaluating logical reasoning independently of other forms of reasoning such as LogiQA \citep{liu2021logiqa} and ReClor \citep{yureclor}; and (2) evaluating only a single type of logic and covering only few logical inference rules as done in FOLIO \citep{han2022folio} and ProntoQA \citep{saparov2023language}. 
Thus, our aim in this work is to address the lacuna of having a more comprehensive set of inference rules for evaluating the logical reasoning ability of LLMs.


To this end, we introduce a systematically created question-answering dataset for the evaluation of logical reasoning ability using a single inference rule, called \textit{LogicBench}. Besides evaluating the logical reasoning ability of LLMs, 
by evaluating models on single inference rules, we can also gain insights into the frequency of text sequences corresponding to these rules in the pre-training data and their impact on model performance.
As illustrated in Figure \ref{fig:logic_type}, \textit{LogicBench} includes a total of 25 reasoning patterns across `propositional, first-order, and non-monotonic' logics.
To the best of the authors' knowledge, this is the first work to study non-monotonic reasoning, as well as various inference rules in propositional and first-order logics including hypothetical and disjunctive syllogism; and bidirectional, constructive, and destructive dilemmas in NLP domain.
To evaluate LLMs using \textit{LogicBench}, we formulate two different tasks: (i) a Binary Question-Answering (BQA) task in which the context comprises logical statements and the models have to determine whether a conclusion given in the question is logically entailed by the context, and (ii) a Multiple-Choice Questions-Answering (MCQA) task where models must select the most appropriate logical conclusion from four distinct options, based on the provided context. 
The rationale behind having BQA and MCQA tasks is that it provides systematic standard metric-based evaluation (i.e., direct comparison of LLMs’ performance in terms of accuracy), which could be more challenging with open-ended question-answer formats.
Examples instances of various reasoning patterns are presented in Table \ref{tab:NL_conversion_examples_small} and App. \ref{app:examples}. 

To construct \textit{LogicBench}, we use a three-stage procedure (refer to \textsection \ref{sec:logicbench}). In the first stage, we prompt GPT-3.5 to generate a variety of coherent natural language sentences having different `ontologies' (i.e., a collection of concepts as car, person, and animals) and their corresponding negations (refer to \textsection \ref{sec:sent_gen}). In the second stage, we generate \textit{(context, question)} pairs where the context represents a natural language narrative consisting of logical statements, and the question is formulated to exhibit the logical conclusion derived from the context. In the third stage, we generate task-specific data instances (i.e., \textit{(context, question, answer)} triplets).

We conduct a comprehensive evaluation with a range of LLMs on \textit{LogicBench} including GPT-4, ChatGPT (GPT-3.5-Turbo), Gemini-Pro, Llama-2-7B-Chat, and Mistral-7B-Instruct using chain-of-thought \citep{wei2022chain}. In particular, we measure the accuracy of LLMs predictions on both BQA and MCQA tasks. 
Our experiments result in several interesting findings such as LLMs often struggle to reason over complex logical contexts that involve only a single inference rule and encounter difficulties with inference rules involving negations.  
Experimental results reveal that these models struggle with respect to many of the inference rules and patterns, suggesting significant room for improvement in their logical reasoning abilities.  
To further demonstrate the use of \textit{LogicBench}, we synthetically augment it and fine-tune T5-large. Our preliminary results (App. \ref{app:discussion_results}) show that this improves the logical reasoning ability of existing models leading to performance improvement on other logic datasets, LogicNLI, and FOLIO ($\sim 2\%$ on an average), and shows competitive performance on LogiQA and ReClor. 

\begin{table}[]
  \centering
  \resizebox{\linewidth}{!}{
    \begin{tabular}{c|c|c|c|c|c}
    \toprule
    \multicolumn{1}{c|}{\multirow{2}[0]{*}{Dataset}} & \multicolumn{3}{c|}{Logic Covered} & \multicolumn{1}{c|}{\multirow{2}[1]{*}{\begin{tabular}[c]{@{}c@{}}Inference Rules/Axioms\\Provided with Data\end{tabular}}}  & \multicolumn{1}{c}{\multirow{2}[1]{*}{\begin{tabular}[c]{@{}c@{}}Generation\\Code Available\end{tabular}}} \\ \cmidrule{2-4} 
    & PL & FL & NM & & \\ \midrule \midrule
    

    Ruletaker & \cellcolor{lightred} \xmark    &   \cellcolor{lightgreen} \checkmark    &   \cellcolor{lightred}\xmark   & \cellcolor{lightred}\xmark   &  \cellcolor{pastelyellow}Human-annotated \\
    
    LogicNLI   &  \cellcolor{lightred}\xmark   &  \cellcolor{lightgreen} \checkmark     &   \cellcolor{lightred}\xmark &  \cellcolor{lightred}\xmark 
    & \cellcolor{pastelyellow}Semi-automated \\ 
    
    ProofWriter     &   \cellcolor{lightgreen} \checkmark    &   \cellcolor{lightgreen} \checkmark    &   \cellcolor{lightred}\xmark   & \cellcolor{lightred}\xmark  & \cellcolor{lightred}\xmark \\
    
    FOLIO     &    \cellcolor{lightred}\xmark   &  \cellcolor{lightgreen} \checkmark     &   \cellcolor{lightred}\xmark   &  \cellcolor{lightred}\xmark  & \cellcolor{pastelyellow}Human-annotated\\
    
    SimpleLogic     &    \cellcolor{lightgreen}\checkmark   &   \cellcolor{lightred}\xmark    &   \cellcolor{lightred}\xmark   &  \cellcolor{lightred}\xmark 
    &   \cellcolor{lightgreen} \checkmark\\
    
    ProntoQA     &  \cellcolor{lightred}\xmark   &  \cellcolor{lightgreen} \checkmark     &   \cellcolor{lightred}\xmark   &  \cellcolor{lightgreen}\checkmark & \cellcolor{lightgreen} \checkmark\\ \midrule
    
    \textbf{LogicBench} &  \cellcolor{lightgreen}\checkmark    & \cellcolor{lightgreen}\checkmark     & \cellcolor{lightgreen}\checkmark     & \cellcolor{lightgreen}\checkmark  & \cellcolor{lightgreen}\checkmark \\ \bottomrule
    \end{tabular}%
    }
    \vspace{-0.2cm}
    \caption{\textit{LogicBench}'s comparison with current datasets}
  \label{tab:dataset_comparison}%
\end{table}

\section{Related Work}

\begin{table*}[htbp]
  \centering
  \resizebox{0.95\linewidth}{!}{
    \begin{tabular}{c|c|c} \toprule
    \textbf{Names} & \textbf{Propositional Logic} & \textbf{Extension to a (restricted) First-order Logic} \\ \midrule
    \begin{tabular}[c]{@{}c@{}}MP\end{tabular}  & $((p \to q) \land p) \vdash q $ & \begin{tabular}[c]{@{}c@{}}$(\forall x(p(x) \to q(x)) \land p(a)) \vdash q(a) $ \end{tabular} \\ \midrule
    
    \begin{tabular}[c]{@{}c@{}}MT\end{tabular} &  $((p\to q)\land \neg q)\vdash \neg p$ &  \begin{tabular}[c]{@{}c@{}}$(\forall x(p(x) \to q(x)) \land \neg q(a)) \vdash \neg p(a) $ \end{tabular} \\ \midrule
    
    \begin{tabular}[c]{@{}c@{}}HS\end{tabular} & $((p\to q))\land (q\to r))\vdash (p\to r)$ & \begin{tabular}[c]{@{}c@{}}$(\forall x((p(x) \to q(x)) \land (q(x) \to r(x))) \vdash (p(a) \to r(a)) $ \end{tabular} \\ \midrule
    
    \begin{tabular}[c]{@{}c@{}}DS\end{tabular} & $((p\lor q)\land \neg p)\vdash q$ & \begin{tabular}[c]{@{}c@{}}$(\forall x(p(x) \lor q(x)) \land \neg p(a)) \vdash q(a) $\end{tabular} \\ \midrule
    
    \begin{tabular}[c]{@{}c@{}}CD\end{tabular} & $((p\to q)\land (r\to s)\land (p\lor r))\vdash (q\lor s)$ & \begin{tabular}[c]{@{}c@{}}$(\forall x((p(x) \to q(x)) \land  (r(x) \to s(x))) \land (p(a) \lor r(a)))\ \vdash (q(a) \lor s(a)) $\end{tabular} \\ \midrule
    
    \begin{tabular}[c]{@{}c@{}}DD\end{tabular} & $((p\to q)\land (r\to s)\land (\neg q\lor \neg s))\vdash (\neg p\lor \neg r)$ & \begin{tabular}[c]{@{}c@{}}$(\forall x((p(x) \to q(x)) \land  (r(x) \to s(x))) \land (\neg q(a) \lor \neg s(a)))\ \vdash (\neg p(a) \lor \neg r(a)) $\end{tabular} \\ \midrule
    
    \begin{tabular}[c]{@{}c@{}}BD\end{tabular} & $((p\to q)\land (r\to s)\land (p\lor \neg s))\vdash (q\lor \neg r)$ & \begin{tabular}[c]{@{}c@{}}$(\forall x((p(x) \to q(x)) \land  (r(x) \to s(x))) \land (p(a) \lor \neg s(a)))\ \vdash (q(a) \lor \neg r(a)) $\end{tabular} \\ \midrule
    
    CT  & $(p\lor q)\vdash (q\lor p)$ & - \\ \midrule
    
    \begin{tabular}[c]{@{}c@{}}MI\end{tabular} &  $(p\to q)\vdash (\neg p\lor q)$ & - \\ \midrule
    
    \begin{tabular}[c]{@{}c@{}}EG\end{tabular} & - &  ${\displaystyle  P\left({a}\right)} \Rightarrow \exists xP\left({x}\right)$ \\ \midrule
    
    \begin{tabular}[c]{@{}c@{}}UI\end{tabular} & - & $\displaystyle \forall x\,A\Rightarrow A\{x\mapsto a\}$ \\ \bottomrule
    
    \end{tabular}%
}
    \caption{Inference rules and (two) axioms that establish the relationship between premises and conclusions. MP: Modus Ponens, MT: Modus Tollens, HS: Hypothetical Syllogism, DS: Disjunctive Syllogism, CD: Constructive Dilemma, DD: Destructive Dilemma, BD: Bidirectional Dilemma, CT: Commutation, MI: Material Implication, EG: Existential Generalization, UI: Universal Instantiation}
    \label{tab:rules}%
    \vspace{-0.2cm}
\end{table*}

As LLMs continue to evolve rapidly, it becomes increasingly crucial to evaluate their diverse reasoning capabilities, as well as those of forthcoming LLMs. LogiQA \citep{liu2021logiqa} and ReClor \citep{yureclor} have made notable contributions by compiling multichoice questions from standardized examinations that demand diverse forms of logical reasoning. In contrast to \textit{LogicBench}, these datasets involve mixed forms of reasoning and do not focus on assessing logical reasoning in isolation. 

A few past attempts have been made to evaluate only logical reasoning while excluding other forms of reasoning. For example, CLUTTER \citep{sinha-etal-2019-clutrr} covers inductive reasoning, \cite{hahn2021teaching} covers temporal logic, and Ruletaker \citep{clark2021transformers} evaluates whether a transformer-based model emulates deductive reasoning over synthetically generated statements in a limited setting. LogicNLI \citep{tian-etal-2021-diagnosing} introduced a diagnostic benchmark for FOL reasoning, with the dataset constructed by automatically generating logic expressions and replacing the entity and attribute placeholders. 

Our proposed dataset is similar (in terms of task formulation) to ProofWriter \citep{tafjord-etal-2021-proofwriter}, FOLIO \citep{han2022folio}, and ProntoQA \citep{saparov2023language} which are QA datasets designed to test reasoning ability. ProofWriter provides multi-hop proofs for each example, while FOLIO gives diverse and complex logical expressions, however, it is only limited to FOL. ProntoQA \citep{saparov2023language} provides explanation and reasoning steps but is limited to modus ponens in FOL. Nevertheless, several crucial attributes motivated us to create \textit{LogicBench} (see Table \ref{tab:dataset_comparison} for comparison). Additional datasets for evaluating logical reasoning also exist such as SimpleLogic \citep{zhang2022paradox} provides a class of logical reasoning problems, TaxiNLI \citep{joshi-etal-2020-taxinli} introduces logical taxonomy in NLI task and RuleBert \citep{saeed-etal-2021-rulebert} covers only soft logical rules. In summary, \textit{LogicBench} evaluates logical reasoning in isolation and provides diverse inference rules and logic types compared to existing datasets. Extended related work is discussed in App. \ref{app:related_work}. 


\section{LogicBench}
\label{sec:logicbench}

\subsection{Logics Types}

\begin{table*}[htbp]
\centering
\resizebox{\linewidth}{!}{
\begin{tabular}{m{11.5cm}|m{11.5cm}}
\toprule
\multicolumn{1}{c|}{\textbf{Basic Default Reasoning}}                                                                                                                                                                                                                                                        & \multicolumn{1}{c}{\textbf{Default Reasoning with Irrelevant Information}}                                                                                                             \\ \midrule
Context: Block A and block B are both heavy objects that are typically found on the table. 
However, there is a possibility that block A might not follow this usual convention. 
It is important to note this exception. \newline \newline Conclusion: B is on the table. 
& Context: In a room filled with various objects, two heavy blocks, block A and block B, stand out. Normally, heavy blocks like these are placed on the table, but surprisingly, block A is not found on the table. On the other hand, block B grabs attention with its vibrant red color. \newline \newline Conclusion: B is on the table.                                                                                        \\ \midrule
\multicolumn{1}{c|}{\textbf{Reasoning about Unknown Expectations}}                & \multicolumn{1}{c}{\textbf{Reasoning about Priorities}}                        \\ \midrule

Context: In this situation, there are three heavy blocks: A, B, and C. Typically, heavy blocks are found on the table. However, it is known that at least one of the blocks, either A or B, is not currently on the table. \newline \newline Conclusion: C is on the table. Exactly one of A, B is not on the table.  & 
Context: John confidently states that the vehicle is situated in the driveway, while Sara adamantly counters, asserting that it is not parked inside the garage. \newline \newline Conclusion: If John's evidence is more reliable than Sara's then the car is parked in the driveway.\\ \bottomrule
\end{tabular}}
\vspace{-0.1cm}
\caption{Illustrative examples of non-monotonic reasoning adapted from \cite{lifschitz1989benchmark}.}
\label{tab:nm_examples}
\end{table*}
    



\paragraph{Propositional Logic (PL)} 
Propositional logic employs a collection of statements or propositions (denoted as $\mathcal{P}={p_1, p_2,..., p_n}$, where $p_i$ represents a proposition) and builds upon them using logical connectives such as `$\land$', `$\lor$', `$\to$', `$\leftrightarrow$', and `$\neg$'. Several inference rules for propositional logic have been defined using which 
given a set of premises, one can derive a sound conclusion. For example, let's consider two propositions: $p_1$, which states "It is raining," and $p_2$, which states "It is cloudy." From these propositions, we can construct a context/knowledge base (KB) with two premises: (1) $p_1 \to p_2$ and (2) $p_1$. With this KB, we can conclude $p_2$. This inference rule is written as $((p_1 \to p_2) \land p_1) \vdash p_2$ and is known as `Modus Ponens'. In our study, we explore nine distinct inference rules of propositional logic, extensions of seven of them with one-variable and a universal quantifier, and two axioms of first-order logic as shown in Table \ref{tab:rules}. These inference rules provide a proper framework for deriving valid conclusions. 

\paragraph{First-order Logic (FOL)} 
In this work, we consider a restricted set of logical axioms for FOL that utilize quantifiers, $\forall$ (universal quantifier) and $\exists$ (existential quantifier). The universal quantifier ($\forall$) denotes that a statement holds true for all instances within a specific category. In contrast, the existential quantifier ($\exists$) indicates that a statement is true for at least one instance within its scope. For instance, a simple extension of propositional `Modus Ponens' is an inference rule where given the premises $\forall(p(x) \rightarrow q(x))$ and $p(a)$, we conclude $q(a)$ (e.g., given ``All kings are greedy” and ``Sam is a king”, we can conclude ``Sam is greedy”). Here, we explore two axioms (EG and UI - in detail in App. \ref{app:FOL_examples}) and various inference rules that incorporate the quantifiers (shown in Table \ref{tab:rules}).


\paragraph{Non-monotonic (NM) Reasoning} 
In this work, we analyze a range of logical reasoning templates in NM logics involving ``Default Reasoning,'' ``Reasoning about Unknown Expectations,'' and ``Reasoning about Priorities.'' These templates are inspired by
the compilation \citep{lifschitz1989benchmark} made in 1989 to evaluate the abilities of various non-monotonic logics that were being developed at that time. 
Below Table \ref{tab:nm_examples} shows examples of NM reasoning. Additional examples are given in App. \ref{app:NM_examples}.

A key aspect of NM logics is to formalize notions such as  ``normally,'' ``typically,'' and ``usually'' that are not directly formalizable using classical quantifiers in the first-order setting. 
The general rule ``Heavy blocks are normally located on the table'' does not imply that ``All heavy blocks are always located on the table''. Rather, this rule allows for exceptions. Our work explores various NM reasoning patterns, as depicted in Figure \ref{fig:logic_type}, to delve deeper into the nuances of this type of reasoning.


\subsection{Data Creation}\label{sub:3.2}

\begin{figure*}[h]
    \centering
    \includegraphics[width=0.75\linewidth]{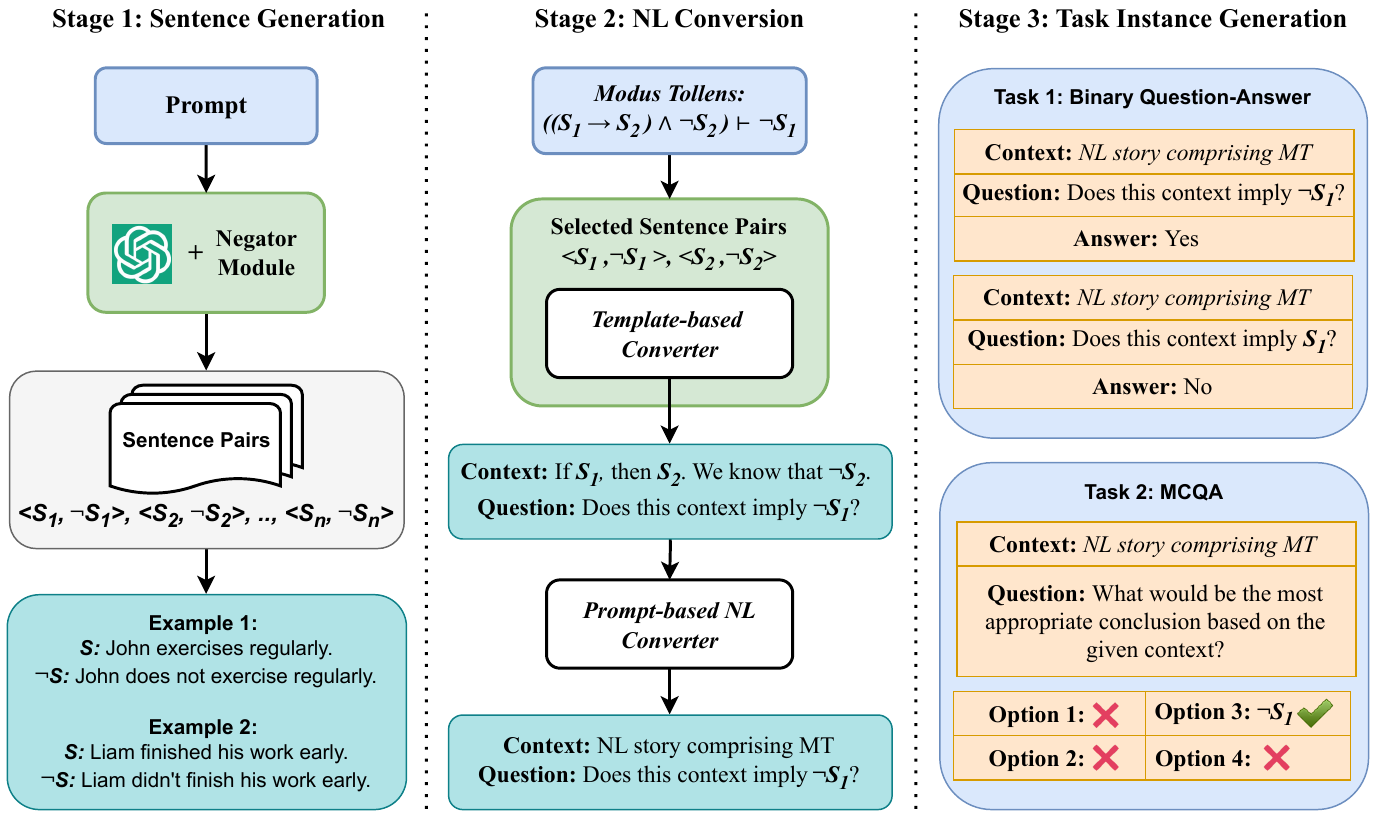}
    \caption{Schematic representation of three-stage procedure for data creation. NL: Natural Language}
    \label{fig:approach}
\end{figure*}

Our data creation procedure, illustrated in Figure \ref{fig:approach}, consists of three stages:

\textbf{1. Sentence Generation:} Starting with a given prompt, we generate coherent sentences and their negations that incorporate different ontologies.

\textbf{2. NL Conversion:} Pairs of \textit{(context, question)} are generated using pre-defined templates from which context is then converted to a natural language narrative using the prompt.

\textbf{3. Task Instance Generation:} Task-specific \textit{(context, question, answer)} triplets are generated. BQA requires answers in the form of ``yes'' or ``no'', whereas MCQA involves selecting one correct option from a set of four. We generate semantically preserving and inverting variations of these triplets to add more diversity for BQA.

Examples of generated data corresponding to each logic type and reasoning patterns are presented in App. \ref{app:examples}.

\subsubsection{Sentence Generation}
\label{sec:sent_gen}

Here, the first step is to generate sentences with diverse \textit{ontologies}. An ontology represents a collection of concepts (e.g. car, person, animals, etc.) along with their corresponding associated properties. To generate these sentences, we prompt the GPT-3.5 model with instructions tailored for each inference rule (more details in App. \ref{app:prompt}). 

An example of a prompt corresponding to the `Modus Tollens' from PL is presented in App. \ref{app:prompt} for better illustration. Note that our objective at this stage is not to generate logical sentences but rather to generate a diverse and coherent set of sentences that encompass various concepts. We also create a negation sentence corresponding to each generated sentence\footnote{We use \url{https://github.com/dmlls/negate} to generate negated sentences}. In this work, the scope of generating negations is simple (refer to App. \ref{app:examples} for examples), however, negations can be more complicated in the case of logic. These generated sentences will be combined with logical connectives in a later stage to form context and questions.



\subsubsection{NL Conversion}
\label{sec:NL_conversion}

Here, the NL conversion is accomplished using two steps. First, we leverage the formal expressions of reasoning patterns to create templates that establish the desired NL formulation for each logical connective (i.e., templatized context). Second, we prompt GPT-3.5 to transform the templatized context into a story/narrative-based context, enhancing its naturalness. For instance, implication: ``$p \to q$'' is expressed as ``If $p$, then $q$'', conjunction: ``$p \land q$'' as ``$p$ and $q$.'', and disjunction: ``$p \lor q$'' as ``At least one of the following is true: (1) $p$ and (2) $q$. Note that we do not know which of (1) and (2) is true. It is possible that only (1) is true, or only (2) is true, or both are true.'' since understanding the logical implication of `or' when integrated into logical formulations posed challenges to both humans and models. With these established formulations, we proceed to utilize the sentences generated in \textsection \ref{sec:sent_gen} to create the templatized context and questions corresponding to reasoning patterns.

Then, the templatized context is converted into a narrative-based context using a prompt-based converter, enhancing its naturalness. The prompt-based converter (essentially, prompting GPT-3.5) ensures that the context is no longer templatized yet follows the logical connection between sentences as mentioned in the logical rule (further details are presented in App. \ref{app:NL_conversion}).  For instance, let's consider the ``Modus Tollens'' from PL ($((p \to q) \land \neg q) \vdash \neg p$), and the ``Bidirectional Dilemma'' from FOL ($\forall x((p(x) \to q(x)) \land (r(x) \to s(x))) \land (p(a) \lor \neg s(a))) \vdash (q(a) \lor \neg r(a))$). For these rules, Table \ref{tab:NL_conversion_examples_small} presents examples of logical templatized and narrative-based context, question, and task instances for both, BQA and MCQA. App. \ref{app:NL_conversion} showcases further examples corresponding to each inference rule and patterns from \textit{LogicBench}.


\begin{table*}[]
\centering
\resizebox{1\linewidth}{!}{

\begin{tabular}{m{6.3cm}|m{10.5cm}|m{7.5cm}|m{8.5cm}}
\toprule
\multicolumn{1}{c|}{Generated Sentences in Stage 1} & \multicolumn{1}{c|}{Context} & \multicolumn{1}{c|}{Binary QA}  & \multicolumn{1}{c}{MCQA} \\ \midrule

Inference rule: \textbf{MT} \newline\newline p: Liam finished his work early.\newline ¬p: Liam did not finish his work early.\newline q: he will order pizza for dinner.\newline ¬q: he will not order pizza for dinner. & \textbf{Templatized Context:} If Liam finishes his work early, then he will order pizza for dinner. He won’t order pizza for dinner.\newline\newline \textbf{NL Context:} Liam knows that if he finishes his work early for the day, he will order pizza for dinner. However, on this particular day, he decided against ordering pizza. & \textbf{Question 1:} Does this imply that Liam didn’t finish his work early? (Yes)\newline \textbf{Question 2:} Does this imply that Liam finishes his work early? (No)  & \textbf{Question:} Based on the context, what conclusion would be deemed most suitable?\newline\newline \textbf{Option 1:} Liam didn't finish his work early. (Yes)\newline \textbf{Option 2:} Sarah had already ordered Chinese takeout. (No)\newline \textbf{Option 3:} Rebecca finished her work early. (No)\newline \textbf{Option 4:} Liam decided to order sushi instead. (No)\\ \midrule

Inference rule: \textbf{BD} \newline \newline p(x): someone drinks lots of water\newline q(x): they will feel hydrated\newline r(x): they eat too much sugar\newline s(x): they will experience a sugar crash\newline p(a): Jane drinks lots of water\newline ¬p(a): Jane does not drink lots of water\newline q(a): she will feel hydrated\newline ¬q(a): she will not feel hydrated\newline r(a): she eats too much sugar\newline ¬r(a): she does not eat too much sugar\newline s(a): she will experience a sugar crash\newline ¬s(a): she will not experience a sugar crash & \textbf{Templatized Context:} If someone drinks lots of water, then they will feel hydrated. If they eat too much sugar, then they will experience a sugar crash. We know that at least one of the following is true (1) Jane drinks lots of water and (2) she won’t experience a sugar crash. Note that we do not know which ones of (1) and (2) are true. It might be the case that only (1) is true, or only (2) is true or both are true.\newline\newline \textbf{NL Context:} If someone consumes a significant amount of water, they will experience a state of hydration. conversely, if excessive amounts of sugar are ingested by them, a sugar crash will ensue. it is known that at least one of the following statements is true: either the Jane consumes ample water or she will not experience a sugar crash. however, the actual veracity of either statement remains ambiguous, as it could be the case that only the first statement is true, only the second statement is true, or both statements are true. & \textbf{Question 1:} Can we say at least one of the following must always be true? (a) she will feel hydrated and (b) she doesn't eat too much sugar (Yes)\newline \textbf{Question 2:} Can we say at least one of the following must always be true? (a) she won't feel hydrated and (b) she eats too much sugar (No)\newline \textbf{Question 3:} Can we say at least one of the following must always be true? (a) she will feel hydrated and (b) she eats too much sugar (No)\newline \textbf{Question 4:} Can we say at least one of the following must always be true? (a) she won't feel hydrated and (b) she doesn't eat too much sugar (No) & \textbf{Question:} Taking into account the context provided, what conclusion would be the most appropriate?\newline\newline \textbf{Option 1:} If Jane consumes ample water, she will experience a sugar crash (No)\newline \textbf{Option 2:} John will feel hydrated or he won't experience a sugar crash (No)\newline \textbf{Option 3:} Jane will feel hydrated or she doesn't eat too much sugar (Yes)\newline \textbf{Option 4:} Jane won't feel hydrated or she will eat too much sugar (No) \\ \bottomrule
\end{tabular}
}
\vspace{-0.2cm}
\caption{Illustrative examples of logical context and questions created using sentences that are generated in the first stage (\textsection \ref{sec:sent_gen}) for both tasks, i.e., BQA and MCQA.}
\label{tab:NL_conversion_examples_small}
\end{table*}

\subsubsection{Task Instance Generation}

After generating the context and questions in \textsection \ref{sec:NL_conversion}, we generate \textit{(context, question, answer)} triplets for both tasks: (i) BQA, and (ii) MCQA. Here, narrative-based context is similar for both BQA and MCQA tasks, only the format of \textit{(question, answer)} pairs are different.

\paragraph{BQA} We generate semantically preserving and inverting variations of questions. Let's consider the example of ``Modus Tollens'' from Table \ref{tab:NL_conversion_examples_small}, having question as: ``Does this imply that Liam didn’t finish his work early?'' In this question, we observe one proposition: $s_1$, representing the statement ``Liam didn't finish his work early,'' can be used to create another question that did not follow the logical rule MT. We can create two possible tuples: $<\neg s_1, yes>, <s_1, no>$. Each tuple has a question-answer combination using proposition $s_1$. Moreover, we do not generate variations for the context since it offers no substantial diversity in the dataset. For question variations, we replace the variation of proposition $<s_1>$ in the original question with the corresponding tuples to add diversity to \textit{LogicBench}. The process allows us to create more variations of the question for BQA tasks, as illustrated in Figure \ref{fig:approach} (Step 3 - Task 1). 

\paragraph{MCQA} A prompt-based approach is used to create different incorrect options. For rule MT, as shown in Table \ref{tab:NL_conversion_examples_small}, option 1 is a correct option that follows the rule MT logically ($<\neg s_1>$), while the other three options are generated using prompting in a way that does not follow the rule being incorrect options. In addition to the options, the question is also replaced by a randomly selected question from the set of five questions. As seen in Figure \ref{fig:approach} (Step 3 - Task 2), there is only one correct option out of four given options. More details related to incorrect option generation and instances examples for both tasks are in the App. \ref{app:examples}, and \ref{E.2}.


\subsection{Statistics and Qualitative Analysis}
\paragraph{Statistics} We introduce two versions of our proposed dataset:\textit{ LogicBench(Eval)}$_{BQA}$ and \textit{LogicBench(Eval)}$_{MCQA}$. Statistics of both versions are presented in Table \ref{tab:stats}. For \textit{ LogicBench(Eval)}$_{BQA}$, out of 1520, 520 samples are for `yes' and 1000 samples are for `no' labels. For \textit{ LogicBench(Eval)}$_{MCQA}$, there are 20 unique samples present for each rule, thus in total 500 unique samples. Furthermore, we synthetically augmented \textit{LogicBench(Eval)} for training purposes (i.e., \textit{LogicBench(Aug)}) which consists of 150 unique data samples for each rule for BQA, resulting in a total of 12908 data samples including variations.
\begin{table}[!h]
\centering
\resizebox{\linewidth}{!}{
\begin{tabular}{c|c|c|c}
\toprule
\textbf{Dataset} & \textbf{\begin{tabular}[c]{@{}c@{}}\# of Instances\\ per Axiom\end{tabular}} & \textbf{\begin{tabular}[c]{@{}c@{}}Total \# of\\ Instances\end{tabular}} & \textbf{\begin{tabular}[c]{@{}c@{}}Total \# of Instances\\ (Including Variations)\end{tabular}} \\ \midrule
\textit{ LogicBench(Eval)}$_{BQA}$ & 20                                                                           & 500                                                                      & 1520                                                                                            \\
\textit{ LogicBench(Eval)}$_{MCQA}$  & 20                                                                          & 500                                                                     & 500                                                                                           \\ \bottomrule
\end{tabular}
}
\vspace{-0.2cm}
\caption{Statistics of \textit{LogicBench(Eval)}}
\vspace{-0.2cm}
\label{tab:stats}
\end{table} 
\paragraph{Data Validation} Throughout the data generation phase of \textit{LogicBench(Eval)}, the authors conduct a review of the logical formations to ensure they follow the intended logical structure. We examine each narrative for any potential discrepancies, ensuring that they are logically sound and correctly represent the intended relationships between propositions. In addition to the logical formation, we also dedicated considerable effort to eliminating typos and validating the grammar. We also analyze the diversity in terms of different ontology and the logical nature of the \textit{LogicBench(Eval)} (presented in App. \ref{app:word_cloud}). We mitigate errors encountered during the validation step (presented in App. \ref{app:error_mitigation}).

\section{Results and Analysis}
\label{sec:results_and_analysis}

\begin{table*}[ht]
  \centering
    \resizebox{\linewidth}{!}{
    \begin{tabular}{c|c|cc|cc|cc|cc|cc} \toprule
      \multirow{2}[0]{*}{\textbf{Type}}    &    \multirow{2}[0]{*}{\textbf{Rules}}  & \multicolumn{2}{c|}{\textbf{Llama-2}} & \multicolumn{2}{c|}{\textbf{Mistral}} & \multicolumn{2}{c|}{\textbf{Gemini}} & \multicolumn{2}{c|}{\textbf{ChatGPT}} & \multicolumn{2}{c}{\textbf{GPT-4}} \\ \cmidrule{3-12}
      &  & $A(No)$ & $A(Yes)$ & $A(No)$ & $A(Yes)$ & $A(No)$ & $A(Yes)$ & $A(No)$ & $A(Yes)$ & $A(No)$ & $A(Yes)$ \\ \midrule
    \multirow{8}[0]{*}{PL} & HS & ~~$100_{0.00}$   & $47.35_{0.03}$  & $98.63_{0.01}$  & $64.24_{0.07}$  & $99.31_{0.01}$  & $59.60_{0.01}$  & ~~$100_{0.00}$  & $66.82_{0.04}$  & ~~$100_{0.00}$  & $84.75_{0.06}$ \\
          & DS & $44.81_{0.05}$  & $56.82_{0.02}$   & $50.77_{0.03}$  & $77.66_{0.05}$  & $68.21_{0.04}$  & $91.65_{0.04}$   & $51.26_{0.04}$  & $80.71_{0.09}$  & $79.96_{0.06}$ & ~~$100_{0.00}$ \\
          & CD & $79.94_{0.03}$    & $25.47_{0.01}$    & $83.37_{0.03}$    & $72.30_{0.12}$    & $85.56_{0.04}$  & $28.15_{0.01}$     & $90.75_{0.06}$  & $38.63_{0.04}$  & $92.85_{0.01}$ & $66.12_{0.05}$ \\
          & DD & $82.22_{0.16}$    & $25.22_{0.01}$    & $71.26_{0.02}$    & $14.16_{0.03}$    & $75.48_{0.04}$  & $25.22_{0.02}$     & $71.28_{0.05}$    & $23.47_{0.03}$  & $84.51_{0.02}$     & $42.40_{0.04}$ \\
          & BD & $48.89_{0.43}$    & $25.13_{0.01}$    & $79.68_{0.01}$    & $47.36_{0.11}$    & $86.81_{0.02}$  & $29.51_{0.01}$     & $84.91_{0.05}$  & $33.47_{0.05}$    & $87.86_{0.04}$ & $59.06_{0.13}$ \\
          & MT & $70.54_{0.03}$  & $71.46_{0.05}$  & $47.96_{0.08}$  & $44.96_{0.10}$  & $75.99_{0.05}$  & $81.47_{0.03}$  & $55.63_{0.02}$  & $66.11_{0.07}$  & $55.28_{0.04}$ & $59.05_{0.07}$ \\
          & MI & $78.57_{0.26}$ & $25.34_{0.01}$  & $75.36_{0.01}$  & $25.51_{0.04}$     & $74.69_{0.06}$  & $24.84_{0.04}$     & $81.60_{0.02}$  & $31.79_{0.03}$  & $91.84_{0.01}$ & $39.72_{0.03}$ \\
          & CT & $70.00_{0.12}$    & $24.98_{0.02}$  & $87.38_{0.01}$  & $71.99_{0.03}$  & $88.31_{0.08}$  & $35.11_{0.07}$  & $89.88_{0.05}$  & $43.33_{0.03}$    & $98.59_{0.01}$ & $60.71_{0.06}$ \\ \midrule
     &   \textbf{Avg}    & $\mathbf{71.87_{0.13}}$ & $\mathbf{37.72_{0.02}}$ & $\mathbf{74.30_{0.03}}$ & $\mathbf{52.27_{0.07}}$ & $\mathbf{81.79_{0.04}}$ & $\mathbf{46.94_{0.03}}$ & $\mathbf{78.16_{0.04}}$ & $\mathbf{48.04_{0.05}}$ & $\mathbf{86.36_{0.03}}$ & $\mathbf{63.98_{0.05}}$ \\  \midrule\midrule
    
    \multirow{9}[0]{*}{FOL} & EG & ~~$100_{0.00}$   & $71.67_{0.03}$   & ~~$100_{0.00}$    & ~~$100_{0.00}$   & $85.68_{0.06}$  & $97.97_{0.04}$   & $96.74_{0.03}$ & $96.75_{0.03}$   & $98.41_{0.03}$ & ~~$100_{0.00}$ \\
          & UI & $80.24_{0.05}$  & $56.04_{0.02}$  & $85.24_{0.04}$  & $85.24_{0.04}$  & $90.69_{0.08}$  & $90.29_{0.04}$  & $85.31_{0.02}$  & $96.19_{0.03}$  & $90.16_{0.002}$ & $91.58_{0.03}$ \\
          & MP & $97.44_{0.04}$  & $73.01_{0.04}$  & $96.28_{0.03}$  & $95.24_{0.05}$  & $98.25_{0.03}$  & $96.97_{0.05}$  & ~~$100_{0.00}$  & $91.03_{0.04}$  & $94.74_{0.00}$ & ~~$100_{0.00}$ \\
          & HS & ~~$100_{0.00}$   & $36.83_{0.01}$  & $98.02_{0.00}$   & $65.00_{0.05}$ & $97.70_{0.00}$  & $52.46_{0.04}$  & $95.14_{0.00}$  & $46.74_{0.04}$  & $89.03_{0.01}$ & $61.21_{0.01}$ \\
          & DS & $58.33_{0.14}$  & $52.08_{0.04}$  & $61.69_{0.03}$  & $92.59_{0.13}$   & $75.96_{0.07}$  & $85.82_{0.07}$  & $72.46_{0.10}$  & $94.71_{0.05}$  & $63.34_{0.04}$ & ~~$100_{0.00}$ \\
          & CD & ~~$100_{0.00}$    & $26.36_{0.01}$    & $82.21_{0.02}$  & $82.02_{0.09}$    & $94.71_{0.02}$  & $33.62_{0.02}$     & $87.87_{0.06}$  & $39.37_{0.06}$  & $92.65_{0.01}$ & $74.72_{0.05}$ \\
          & DD & $54.37_{0.19}$    & $23.36_{0.01}$    & $73.78_{0.01}$  & ~~$6.67_{0.11}$    & $82.70_{0.02}$ & $28.96_{0.01}$     & $72.70_{0.08}$  & $26.42_{0.11}$  & $85.54_{0.04}$ & $54.55_{0.08}$ \\
          & BD & $92.31_{0.13}$    & $25.78_{0.01}$    & $75.76_{0.01}$  & $44.44_{0.10}$    & $82.82_{0.08}$   & $28.93_{0.05}$  & $85.50_{0.03}$  & $38.82_{0.02}$  & $84.22_{0.02}$ & $66.67_{0.10}$ \\
          & MT & $77.11_{0.05}$  & $88.41_{0.09}$  & $69.01_{0.06}$  & $91.71_{0.01}$ & $71.18_{0.10}$  & $89.58_{0.04}$  & $62.46_{0.03}$  & $92.80_{0.06}$ & $64.69_{0.04}$ & ~~$100_{0.00}$ \\ \midrule
     &   \textbf{Avg}    & $\mathbf{84.42_{0.07}}$ & $\mathbf{50.39_{0.03}}$ & $\mathbf{82.44_{0.02}}$ & $\mathbf{73.66_{0.06}}$ & $\mathbf{86.63_{0.05}}$ & $\mathbf{67.17_{0.04}}$ & $\mathbf{84.24_{0.04}}$ & $\mathbf{69.20_{0.05}}$ & $\mathbf{84.75_{0.02}}$ & $\mathbf{83.19_{0.02}}$ \\ \midrule\midrule
    
    \multirow{8}[0]{*}{NM} & DRI & $29.06_{0.05}$  & $40.98_{0.04}$  & $61.51_{0.05}$  & $72.22_{0.05}$  & $65.45_{0.03}$  & $78.50_{0.02}$  & $55.77_{0.00}$  & $88.89_{0.09}$   & $81.29_{0.05}$ & ~~$100_{0.00}$ \\
          & DRS & $66.56_{0.03}$  & $18.40_{0.01}$   & $70.13_{0.01}$    & ~~$6.20_{0.01}$   & $68.05_{0.01}$  & ~~$7.20_{0.06}$   & $69.06_{0.01}$  & ~~$0.00_{0.00}$  & $77.59_{0.02}$ & $39.32_{0.06}$ \\
          & DRD & $49.21_{0.06}$ & $50.47_{0.04}$    & $77.64_{0.02}$  & $97.78_{0.04}$  & $68.10_{0.05}$  & $97.43_{0.04}$  &  $55.11_{0.02}$  & ~~$100_{0.00}$  & $86.96_{0.00}$ & ~~$100_{0.00}$ \\
          & DRO & $53.25_{0.01}$  & $53.60_{0.02}$  & $68.30_{0.04}$  & $89.56_{0.11}$  & $57.85_{0.04}$  & $74.03_{0.08}$  & $50.85_{0.01}$  & $66.67_{0.58}$   & $55.56_{0.00}$ & ~~$100_{0.00}$ \\
          & RE1 & $94.07_{0.06}$  & $27.55_{0.01}$  & $78.79_{0.02}$  & $44.39_{0.03}$    & $84.04_{0.04}$  & $41.98_{0.12}$  &  $75.82_{0.00}$  & $31.74_{0.03}$  & $85.44_{0.02}$ & ~~$100_{0.00}$ \\
          & RE2 & $54.17_{0.47}$   & $53.62_{0.05}$   & $81.31_{0.04}$  & $85.67_{0.03}$  & $58.48_{0.03}$  & $82.22_{0.17}$  &  $64.67_{0.11}$ & $60.68_{0.02}$ & $59.42_{0.00}$ & ~~$100_{0.00}$ \\
          & RE3 & $39.05_{0.03}$  & $38.05_{0.10}$    & $78.61_{0.04}$  & $83.59_{0.01}$  & $64.67_{0.08}$  & $79.68_{0.02}$    &  $59.05_{0.04}$ & $83.57_{0.08}$ & $83.12_{0.02}$ & $89.08_{0.01}$ \\
          & RAP & $76.82_{0.05}$  & $67.22_{0.03}$  & $51.73_{0.02}$  & $60.48_{0.11}$  & $75.58_{0.04}$  & $96.67_{0.06}$    & $61.44_{0.02}$  & $92.16_{0.03}$  & $66.07_{0.05}$ & $98.72_{0.02}$ \\ \midrule
      &  \textbf{Avg}     & $\mathbf{57.78_{0.09}}$ & $\mathbf{43.74_{0.04}}$ & $\mathbf{71.00_{0.03}}$ & $\mathbf{67.49_{0.05}}$ & $\mathbf{67.78_{0.04}}$ & $\mathbf{69.71_{0.07}}$ & $\mathbf{61.47_{0.03}}$ & $\mathbf{65.46_{0.10}}$ & $\mathbf{74.43_{0.02}}$ & $\mathbf{84.75_{0.02}}$ \\    \bottomrule
    \end{tabular}%
    }
    \caption{Evaluation of LLMs in terms of label-wise accuracy on \textit{LogicBench(Eval)}$_{BQA}$, where $A(Yes)$ and $A(No)$ denote the accuracy for the $Yes$ and $No$ labels, respectively. DRI: Default Reasoning with Irrelevant Information, DRS: Default Reasoning with Several Defaults, DRD: Default Reasoning with a Disabled Default, DRO: Default Reasoning in an Open Domain, RE1: Reasoning about Unknown Expectations I, RE2: Reasoning about Unknown Expectations II, RE3: Reasoning about Unknown Expectations III, RAP: Reasoning about Priorities}
    \label{tab:main_result}%
\end{table*}%

\subsection{Experimental Setup}
\label{sec:experiments}


\paragraph{Task Formulation} For BQA, let us consider a set of data instances $\mathcal{I}_{r, L}$ corresponding to the inference rule $r$ and logic type $L$. In this set, $i^{th}$ instance is represented as $\mathcal{I}^i_{r, L} = \{(c_i, Q_i)\}$ where $c_i$ represents narrative context and $Q_i = \{q_1, q_2, ..., q_n\}$ represents set of question and its variations corresponding to $i^{th}$ instance. As discussed in \textsection \ref{sec:logicbench}, each context ($c$) represents logical rules (e.g., All cats have fur. Tom is a cat.) and question ($q$) represents the conclusion (e.g., Does Tom have fur?). To each context and question pair, i.e., $<c,q>$, we assign a label from the set $\mathcal{Y} = \{Yes, No\}$. We assign a label $Yes$ if the conclusion logically entails the context, otherwise, assign a label $No$. To evaluate any LLMs on this setup, we provide $<p,c,q>$ as input to predict a label from $\mathcal{Y}$ where $p$ is a natural language prompt. In the set $\mathcal{I}_{r, L}$ for MCQA, $i^{th}$ instance is represented as $\mathcal{I}^i_{r, L} = \{(c_i, q_i, O_i)\}$ where $c_i$ represents narrative context and $q_i$ represents question and $O_i = \{o_1, o_2, o_3, o_4\}$ represents four option choices. To each context and question pair, i.e., $<c,q>$, we assign a label from the set $\mathcal{Y} = \{o_1, o_2, o_3, o_4\}$. We assign a label $o_1$ if the correct conclusion is presented in the first option, and likewise for other labels. To evaluate any LLMs on this setup, we provide $<p,c,q,o>$ as input to predict a label from $\mathcal{Y}$.


\paragraph{Experiments} 
We evaluate a range of prompting models including GPT-4, ChatGPT (GPT-3.5-Turbo), Google Gemini-Pro, Llama-2-7B-Chat, and Mistral-7B-Instruct-v0.2. Each model is evaluated in a zero-shot setting where the chain-of-thought prompt is provided to the model without any in-context examples. This approach allows us to determine LLM's inherent ability to do logical reasoning (based on pre-training), as we can not expect that various logical inference rules/patterns will always be made part of prompts.  However, we do evaluate these models in a few-shot setting, and present the results in App. \ref{app:few_shot}. 





\paragraph{Metrics} Here, we evaluate performance in terms of accuracy for both tasks, BQA and MCQA. For the BQA, we measure accuracy corresponding to each label, i.e., $A(Yes)$ and $A(No)$.
We evaluate each model on three different chain-of-thought prompts and report average results across these prompts. All prompts used for experiments are described in App. \ref{app:exp_prompts}.


\begin{table}[h]
\centering
\resizebox{\linewidth}{!}{
\begin{tabular}{c|c|c|c|c|c|c}
\toprule
\textbf{Type}        & \textbf{Rules} & \textbf{Llama-2} & \textbf{Mistral} & \textbf{Gemini} & \textbf{ChatGPT} & \textbf{GPT-4} \\ \midrule
\multirow{8}{*}{PL}  & HS             &  $86.67_{0.08}$ & $93.33_{0.06}$ & ~~$100_{0.00}$ & $91.67_{0.03}$        & ~~$100_{0.00}$               \\  
                     & DS             &   $63.33_{0.12}$ & $60.00_{0.09}$ & $86.67_{0.10}$ & $96.67_{0.06}$             &  $95.00_{0.00}$              \\  
                     & CD             &   $80.00_{0.05}$ & $70.00_{0.13}$ & $96.67_{0.03}$ & $90.00_{0.00}$                &  ~~$100_{0.00}$              \\  
                     & DD             &  $43.33_{0.03}$ & $30.00_{0.10}$ & $90.00_{0.05}$ & $73.33_{0.06}$                  &  $88.33_{0.03}$              \\  
                     & BD             &   $51.67_{0.03}$ & $53.33_{0.03}$ & $86.67_{0.03}$ & $68.33_{0.06}$             &   $83.33_{0.03}$             \\ 
                     & MT             &  $31.67_{0.03}$ & $60.00_{0.05}$ & $78.33_{0.08}$ & $73.33_{0.08}$                &  $76.67_{0.03}$              \\
                     & MI             &  $33.33_{0.13}$ & $35.00_{0.05}$ & $71.67_{0.06}$ & $63.33_{0.08}$       &   $73.33_{0.14}$             \\ 
                     & CT             &  $73.33_{0.10}$ & $68.33_{0.08}$ & ~~$100_{0.00}$ & $98.33_{0.03}$              &   ~~$100_{0.00}$             \\ \midrule
                     & \textbf{Avg}   & $\mathbf{57.92_{0.07}}$ & $\mathbf{58.75_{0.07}}$ & $\mathbf{88.75_{0.04}}$ & $\mathbf{81.88_{0.05}}$  & $\mathbf{89.58_{0.03}}$                 \\ \midrule \midrule
\multirow{9}{*}{FOL} & EI & $80.00_{0.00}$ & $85.00_{0.05}$ & $95.00_{0.00}$ & $93.33_{0.03}$ &   ~~$100_{0.00}$             \\
                    & UI & $63.33_{0.03}$ & $75.00_{0.05}$ & $98.33_{0.03}$ & $91.67_{0.03}$ &   $98.33_{0.03}$             \\
                    & MP & $85.00_{0.09}$ & $98.33_{0.03}$ & ~~$100_{0.00}$ & ~~$100_{0.00}$ &   ~~$100_{0.00}$             \\
                    & HS & $61.67_{0.06}$ & $70.00_{0.05}$ & $81.67_{0.03}$ & $73.33_{0.06}$ &   $76.67_{0.03}$             \\
                    & DS & $43.33_{0.06}$ & $36.67_{0.03}$ & $70.00_{0.05}$ & $78.33_{0.10}$ &  $95.00_{0.05}$              \\
                    & CD & $75.00_{0.05}$ & $61.67_{0.06}$ & $93.33_{0.03}$ & $80.00_{0.05}$ &   $91.67_{0.03}$             \\
                    & DD & $36.67_{0.06}$ & $46.67_{0.03}$ & $85.00_{0.05}$ & $71.67_{0.06}$ &   $93.33_{0.03}$             \\
                    & BD & $35.00_{0.05}$ & $43.33_{0.06}$ & $78.33_{0.06}$ & $66.67_{0.13}$ &  $91.67_{0.06}$              \\
                    & MT & $41.67_{0.03}$ & $66.67_{0.03}$ & $81.67_{0.06}$ & $86.67_{0.06}$ &   $86.67_{0.03}$             \\ \midrule
                     & \textbf{Avg}   &     $\mathbf{57.96_{0.05}}$ & $\mathbf{64.81_{0.04}}$ & $\mathbf{87.04_{0.03}}$ & $\mathbf{82.41_{0.06}}$         &   $\mathbf{91.51_{0.04}}$             \\ \midrule \midrule
\multirow{8}{*}{NM}  & DRI            &$38.33_{0.03}$ & $28.33_{0.06}$ & $58.33_{0.08}$ & $66.67_{0.06}$  &  $90.00_{0.00}$              \\  
                     & DRS            &$41.67_{0.08}$ & $16.67_{0.10}$ & $45.00_{0.10}$ & $41.67_{0.10}$ &  $55.00_{0.05}$              \\  
                     & DRD            & $55.00_{0.00}$ & $50.00_{0.05}$ & $48.33_{0.03}$ & $71.67_{0.10}$ &  $80.00_{0.05}$              \\  
                     & DRO            & $21.67_{0.03}$ & $21.67_{0.03}$ & $53.33_{0.03}$ & $38.33_{0.08}$ &  $45.00_{0.00}$              \\  
                     & RE1            &  $51.67_{0.03}$ & $31.67_{0.08}$ & $70.00_{0.00}$ & $65.00_{0.05}$   & $95.00_{0.05}$               \\  
                     & RE2            &   $65.00_{0.05}$ & $75.00_{0.00}$ & $68.33_{0.08}$ & $61.67_{0.03}$    & $66.67_{0.06}$               \\  
                     & RE3            &  $31.67_{0.03}$ & $33.33_{0.03}$ & $61.67_{0.08}$ & $70.00_{0.05}$    &  $68.33_{0.03}$              \\  
                     & RAP            &   $46.67_{0.08}$ & $35.00_{0.09}$ & $33.33_{0.03}$ & $55.00_{0.05}$       &  $51.67_{0.03}$              \\ \midrule
                     & \textbf{Avg}            &  $\mathbf{43.96_{0.04}}$ & $\mathbf{36.46_{0.05}}$ & $\mathbf{54.79_{0.05}}$ & $\mathbf{58.75_{0.07}}$ &  $\mathbf{68.96_{0.03}}$              \\ \bottomrule
\end{tabular}
}
\caption{Evaluation of LLMs in terms of accuracy on \textit{LogicBench(Eval)}$_{MCQA}$.}
\label{tab:mcq_result}%
\end{table}

\subsection{Main Results}

Table \ref{tab:main_result} and Table \ref{tab:mcq_result} represent inference rule-wise performance corresponding to each LLMs for the BCQ and MCQA tasks, respectively. Specifically, Table \ref{tab:main_result} provides label-wise accuracy ($A(Yes)$ and $A(No)$) for the BQA task, and Table \ref{tab:mcq_result} provides overall accuracy for the MCQA task. Both tables provide valuable insights into the performance of different models on various logic types and lead to several interesting findings. From Table \ref{tab:main_result}, we can observe that ChatGPT achieves 48.04\%, and GPT-4 shows a performance of 63.98\% $A(Yes)$ on average which indicates the challenge of classical logical reasoning (PL) even for larger LLMs such as ChatGPT and GPT-4. Furthermore, we can observe that models struggle more with inference rules of PL compared to FOL and NM. In addition, it is noticeable that each model performs relatively better on questions with a negative response (i.e., $No$) compared to questions with a positive response (i.e., $Yes$). This observation suggests that the models struggle to fully comprehend the logical relationship between the context and the conclusion (i.e., lower $A(Yes)$). However, they demonstrate a relatively stronger understanding when the relationship is contradictory in nature (i.e., higher $A(No)$). From Table \ref{tab:mcq_result}, we can observe that larger models exhibit superior performance in selecting the correct choice to arrive at a logical conclusion. Interestingly, the performance of models decreases when the inference rules are longer or include negations. In contrast to Table \ref{tab:main_result}, for MCQA, LLMs show superior performance for PL and FOL compared to NM. To further investigate these findings and provide a detailed analysis, we perform a thorough study of reasoning chains generated by LLMs and present our insights in the subsequent section.

\subsection{Analysis and Discussion}

\paragraph{Human Performance}
We conduct a human evaluation on a subset of LogicBench(Eval) for both tasks, BQA and MCQA. Specifically, we selected 50 unique instances covering all 25 reasoning patterns from LogicBench(Eval). This selection resulted in total instances of 153 <context, question> pairs for BQA, and 50 <context, question, choices> pairs for MCQA. We hired three graduate student volunteers to provide the evaluations. The task instructions given to all three annotators closely resemble the prompts provided to models (App. \ref{app:exp_prompts}). Each instance pair is answered/annotated by three different annotators with 0.785 inter-annotator agreement (measured with raw/observed agreement) for BQA and 0.813 for MCQA.

From the results (Table \ref{tab:human_eval}) for BQA, we see that humans achieve more than $\sim85\%$ accuracy on various logic types on LogicBench(Eval) which indicates the capability of humans to comprehend single-step logical reasoning effectively. From Table \ref{tab:main_result}, we observe that the average performance of all models is below human performance indicating room for improvement in their reasoning capabilities. From Table \ref{tab:mcq_result}, we make similar observations for MCQA. However, we can see that the performance of NM for the MCQA task remains a challenge for both humans and LLMs.

\begin{table}[]
\centering
\resizebox{0.8\linewidth}{!}{
\begin{tabular}{c|c|c|c}
\toprule
                                      & \multicolumn{2}{c|}{\textbf{BQA}}                     & \textbf{MCQA}     \\ \cmidrule{2-4} 
\multirow{-2}{*}{\textbf{Logic Type}} & \multicolumn{1}{c|}{\textbf{A(No)}} & \textbf{A(Yes)} & \textbf{Accuracy} \\ \midrule
\textbf{PL}                                                   & \multicolumn{1}{c|}{85.42\%}        & 84.17\%         & 100\%             \\
\textbf{FOL}                                                  & \multicolumn{1}{c|}{90.74\%}        & 91.18\%         & 97.03\%           \\
\textbf{NM}                                                   & \multicolumn{1}{c|}{72.22\%}        & 88.46\%         & 62.05\%           \\ \midrule
\textbf{Avg.}                                                 & \multicolumn{1}{c|}{82.79\%}        & 87.94\%         & 86.36\%           \\ \bottomrule
\end{tabular}
}
\caption{Human performance on three logic types averaged across three annotators for both tasks.}
\label{tab:human_eval}%
\end{table}








\paragraph{Lower performance of LLMs on PL as compared to NM and FOL for BQA.}

In the development of AI, NM logic was partly developed to formalize natural language constructs, such as “normally birds fly’’, that were not formalizable in a straightforward manner using classical mathematical logics. Thus, while it was difficult for researchers to come up with non-monotonic logics and formalize non-monotonic reasoning, the fact that they were usually motivated by natural language examples, suggests that many of the non-monotonic reasoning aspects are present in the NL text in the wild that is used in the pre-training of the ultra-large LLMs such as GPT4. While from human experience and complexity theory, FOL is harder than PL in general; in the LLM context, the crucial factor becomes what kind of logical sentences LLMs are pre-trained on. It seems that LLMs are pre-trained more on simple FOL sentences than on simple PL sentences (see Appendix \ref{app:discussion_results} for further discussion). On the other hand, some PL features are perhaps less prevalent in human writing (on which LLMs are pre-trained) - such as Modes Tollens. Table \ref{tab:main_result} shows that GPT-4 achieves $\sim 85\%$ accuracy ($A(Yes)$) for simple inference rules such as HS(PL). However, GPT-4 performance dropped to $\sim 59\%$ $A(Yes)$ for PL(MT).


\paragraph{Negations are hard to understand when embedded with logical rules.} Regarding PL and FOL, it is apparent that the models struggle more with the DD and MT inference rules. A closer look at Table \ref{tab:rules} reveals that all of these rules include examples where the models need to draw conclusions based on negated premises. This indicates that the models encounter difficulties when negated premises are introduced. We also analyze the effect of negations on the reasoning chain (see App. \ref{app:discussion_results}).   


\paragraph{Longer inference rules are still challenging.} Table \ref{tab:main_result} indicates that the models face challenges when handling longer rules, such as BD, CD, and DD, both in PL and FOL. Hence, it can be concluded that these models struggle with longer logical dependencies in the premise, particularly when a higher number of propositions are present. In the case of NM reasoning, the models exhibit lower performance in DRS, indicating that a higher number of premises often leads to more frequent mistakes. 

\paragraph{LLMs sometimes overlook contextual information.} We investigate the LLMs’ logical reasoning ability in natural language, not in artificial logical formulations. Hence, we note that LLMs sometimes hallucinate information and overlook contextual information, leading to incorrect conclusions. To analyze this, we manually examine all the reasoning chains generated for instances sharing the same contexts in both BQA and MCQA tasks. We observe that, although this pattern is not dominant, it affects BQA more than MCQA. For a more in-depth analysis, please refer to App. \ref{app:discussion_results}.



\paragraph{Large models are better logical reasoners.} We analyze the results of both smaller (Llama-2-7B and Mistral-7B) and larger (ChatGPT, GPT-4, and Gemini) models. Table \ref{tab:main_result} and Table \ref{tab:mcq_result} show that larger models tend to exhibit higher performance across different types of logic. We further investigate an additional model with an intermediate size: Yi-34B-chat \cite{young2024yi} (results are presented in App. \ref{app:discussion_results}). When compared to the Llama-2-7B, the Yi-34B model (5x larger than Llama-7B) shows improvement in average performance across three logic types. Similarly, GPT-4 outperforms Yi-34B. This suggests that increasing the model size leads to substantial gains in performance, indicating the influence of larger model capacities on carrying out better logical reasoning.



\paragraph{Performance of BQA \textit{vs.} MCQA}
From Table \ref{tab:main_result} and Table \ref{tab:mcq_result}, we can see the overall performance of LLMs is higher on $PL_{MCQA}$ compared to $PL_{BQA}$. Conversely, the performance is lower on $NM_{MCQA}$ compared to $NM_{BQA}$. For PL, the performance gaps between the CT and DD inference rules primarily contributed to this trend, and DRO, RAP, and DRD for NM. We analyze the reasoning chains associated with these inference rules and presented our detailed observations in App. \ref{app:discussion_results}.  


\paragraph{Effect on other logic datasets}
We trained the T5-large model on the \textit{ LogicBench(Aug)} resulting in a model named LogicT5. Furthermore, we performed fine-tuning on four other logical reasoning datasets: LogiQA, Reclor, LogicNLI, and FOLIO. Further discussion is presented in App. \ref{app:discussion_results}.

\section{Conclusions}
\label{sec:conclusion}

In this work, we evaluated the logical reasoning ability of LLMs on 25 distinct inference rules and reasoning patterns covering PL, FOL, and NM logics. To this end, we introduced \textit{LogicBench}, a natural language question-answering dataset focusing on evaluating a single inference rule. We devised two tasks using \textit{LogicBench}: (i) BQA, and (ii) MCQA. We evaluated a range of LLMs including GPT-4, ChatGPT, Gemini-Pro, Llama-2, and Mistral on both tasks. Experimental results showed that LLMs do not perform well on \textit{LogicBench}, even though they require the application of only a single inference rule. Furthermore, we also augmented \textit{LogicBench} to \textit{LogicBench(Aug)}, which can be utilized for training purposes. Using  \textit{LogicBench(Aug)}, we demonstrated that LLMs trained using it showcase an improved understanding of logical reasoning, resulting in a better performance on existing logic datasets.

\section*{Limitations}

While \textit{LogicBench} encompasses 25 distinct inference rules spanning three logic types (significantly more than any previous study) to comprehensively evaluate the logical reasoning capabilities of LLMs, it can be further extended by incorporating additional inference rules and logic types.
However, with respect to first-order logic and logics with quantified variables, there can be an infinite number of such rules.  In this study, we focused solely on evaluating model performance using a single inference rule; however, an interesting future direction can be enhancing the depth of reasoning complexity (i.e., multi-step reasoning) by incorporating combinations of inference rules to derive conclusions. We also note that this research is limited to the English language and can be extended to multilingual scenarios for evaluating the logical reasoning ability of LLMs.


\section*{Ethics Statement}

We have used AI assistants (Grammarly and ChatGPT) to address the grammatical errors and rephrase the sentences.

\section*{Acknowledgement}
We thank the anonymous reviewers for constructive suggestions, and the computer science graduate students of Arizona State University (ASU) who helped with the human annotations. We extend our gratitude to the Research Computing (RC) at ASU for providing computing resources for experiments. We acknowledge support by a 2023 Spring Amazon Research Award (ARA), and an award by Cisco via Silicon Valley Foundation.

\bibliography{custom}

\clearpage

\appendix

\section{Example Prompt for Sentence Generation} 
\label{app:prompt}

Below is the general prompt structure prompted to generate data. The prompt schema, as depicted in Figure \ref{fig:prompt}, comprise three crucial components:
\begin{figure}[!h]
  \begin{center}
    \includegraphics[width=0.7\linewidth]{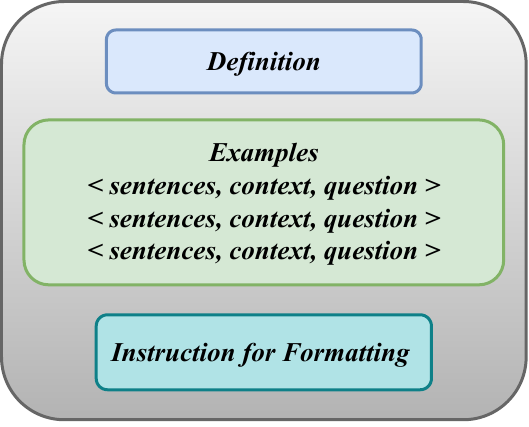}
  \end{center}
  \caption{Schematic representation of prompt.}
  \label{fig:prompt}
\end{figure}

\textit{\textbf{Definition}} provides a detailed explanation of the task and offers a natural language representation of the reasoning pattern for which we are generating sentences.

\textit{\textbf{Examples}} provide sample sentences that need to be generated. We also illustrate how these sentences will be utilized in later stages, emphasizing the importance of coherence and the inclusion of relevant ontological concepts.

\textit{\textbf{Format}} We provide specific formatting instructions to guide the generation of sentences.

\begin{figure}[h]
  \begin{center}
    \includegraphics[width=0.9\linewidth]{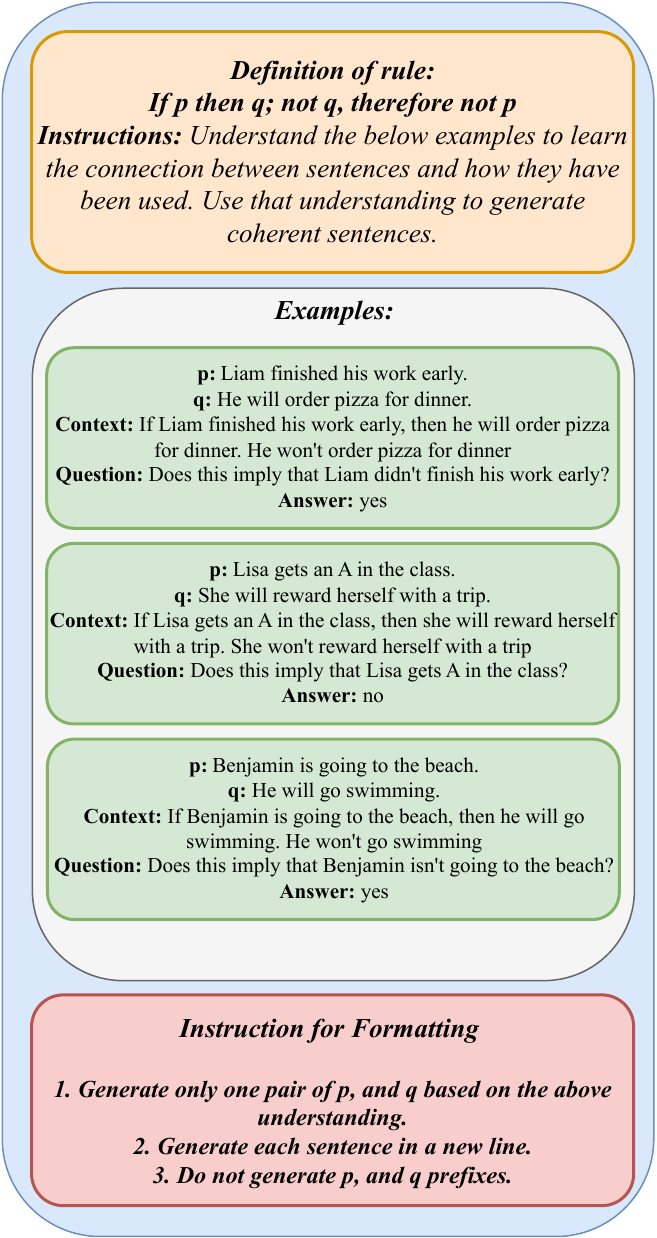}
  \end{center}
  \caption{Example prompt for \textit{Modus Tollens} inference rule from PL.}
  \label{fig:extend_prompt}
\end{figure}

Figure \ref{fig:extend_prompt} illustrates an example prompt for the inference rule, namely, `modus tollens' from propositional logic (PL). Modus tollens is formally represented as $((p\to q)\land \neg q)\vdash \neg p$, which can be understood in natural language as ``If $p$ implies $q$, and we know $\neg q$, then we can conclude $\neg p$.'' In this prompt, the definition provides a comprehensive description of the inference rule in natural language. To encourage the generation of more relevant and coherent sentences, the prompt includes an examples section that demonstrates how the generated sentences will be utilized in a later stage. This serves, as an illustration, to guide GPT-3 in producing suitable outputs. In Figure \ref{fig:extend_prompt}, we present three examples involving sentences $p$ and $q$, along with their respective contexts and questions. The prompt also includes instructions on how the generated sentences should be formatted.

\section{Extended Related Work} \label{app:related_work}

As LLMs such as GPT-4, and Bard continue to evolve rapidly, it becomes increasingly crucial to evaluate their diverse language capabilities, as well as those of forthcoming LLMs.
Recently, many datasets have been created that evaluate different language understanding skills such as pronoun resolution \citep{sakaguchi2021winogrande,levesque_winograd_2012}, commonsense reasoning \citep{talmor-etal-2019-commonsenseqa}, numerical reasoning \citep{dua-etal-2019-drop,patel-etal-2021-nlp,mishra-etal-2022-numglue}, qualitative reasoning \citep{tafjord-etal-2019-quartz, tafjord2019quarel}, temporal reasoning \citep{zhou-etal-2019-going}, and feasibility reasoning \citep{gupta2022john}. Now, we present the advancements in prompt and instruction tuning using LLMs.

\paragraph{Prompt Learning} The introduction of LLMs has significantly shifted the research trend in NLP to prompt-based learning methodologies \citep{liu2021pre}. Many studies have been conducted to investigate the efficacy of prompt-based learning in various applications including Text classification \citep{yin2019benchmarking}, Natural Language Inference (NLI) \citep{schick2020exploiting}, and Question Answering (QA) \citep{jiang2020can}, Information Extraction (IE) \citep{chen2021adaprompt, cui2021template}, to name a few. In a recent development, the T0 model employs prompts to achieve zero-shot generalization across various NLP tasks \citep{sanh2021multitask}. \citet{le2021many} suggested that the use of prompts could be as valuable as hundreds of data points on average. 



\paragraph{Instruction Learning} \citet{efrat2020turking} was focused on whether existing LLMs understand instructions. The same work in the field of instruction by \citet{hase2021can, ye2021zero, gupta2021towards, zhong2021adapting} has been proposed to show that models follow natural language instructions. In addition, \citet{weller2020learning} developed a framework focusing on NLP systems that solve challenging new tasks based on their description. \citet{mishra2021cross} have proposed natural language instructions for cross-task generalization of LLMs. Similarly, PromptSource \citep{sanh2021multitask} and FLAN \citep{wei2021finetuned} were built for leveraging instructions and achieving zero-shot generalization on unseen tasks. Moreover, \citet{parmar-etal-2022-boxbart} shows the effectiveness of instructions in multi-task settings for the biomedical/clinical domain. Furthermore, \citet{mishra2021reframing} discussed the impact of task instruction reframing. \citet{min2021metaicl} introduced a framework to better understand in-context learning. \citet{ouyang2022training} proposed the InstructGPT model that is fine-tuned with human feedback to follow instructions. \citet{wang2022instructionner} has developed an instruction-based multi-task framework for few-shot Named Entity Recognition (NER) tasks. In addition, many approaches have been proposed to improve model performance using instructions \citep{wu2022ai, lin2021few, wang-etal-2022-super, luo2022biotabqa, kuznia-etal-2022-less, patel-etal-2022-question, mishra2022help}.

\paragraph{Logic and NLI Datasets} FraCas \citep{FRACAS} offers a unique approach to temporal semantics by converting syntax trees into logical formulas tailored for inference, emphasizing temporal elements such as references, adverbs, aspectual classes, and progressives. The Monotonicity Entailment Dataset (MED) \citep{MED} dives deep into monotonicity reasoning within NLI, probing the synergy between lexical and syntactic structures and spotlighting inherent challenges in both upward and downward monotonic reasoning trajectories. The SICK \citep{SICK} dataset, with its foundation in 10,000 English sentence pairs, is designed to rigorously evaluate semantic relatedness and entailment, leveraging crowdsourced annotations for precision. HANS, or Heuristic Analysis for NLI Systems \citep{HANS}, stands out by rigorously scrutinizing the dependability of NLI models, putting the spotlight on potential pitfalls tied to syntactic heuristics such as lexical overlap. Lastly, CAD \citep{CAD} introduces a meticulously crafted dataset from Reddit entries, targeting the detection of online abuse. \citet{nakamura-etal-2023-logicattack} introduced LogicAttack, a method for performing adversarial attacks on NLI models using PL inference rules and proposed dataset of $\sim9k$ attack samples derived from the SNLI dataset. In contrast to these works, \textit{LogicBench} evaluates the logical reasoning capabilities of LLMs beyond the NLI task, focusing on the application of individual inference rules.


\section{Examples of Data Instances} 
\label{app:examples}

This section provides examples of \textit{(context, question, answer)} tuples corresponding to each inference rule and reasoning pattern. Additionally, it highlights the diverse range of question variations within the dataset associated with each inference rule and reasoning pattern. 

\subsection{Word Cloud}
\label{app:word_cloud}

\begin{figure}
  \begin{center}
    \includegraphics[width=0.9\linewidth]{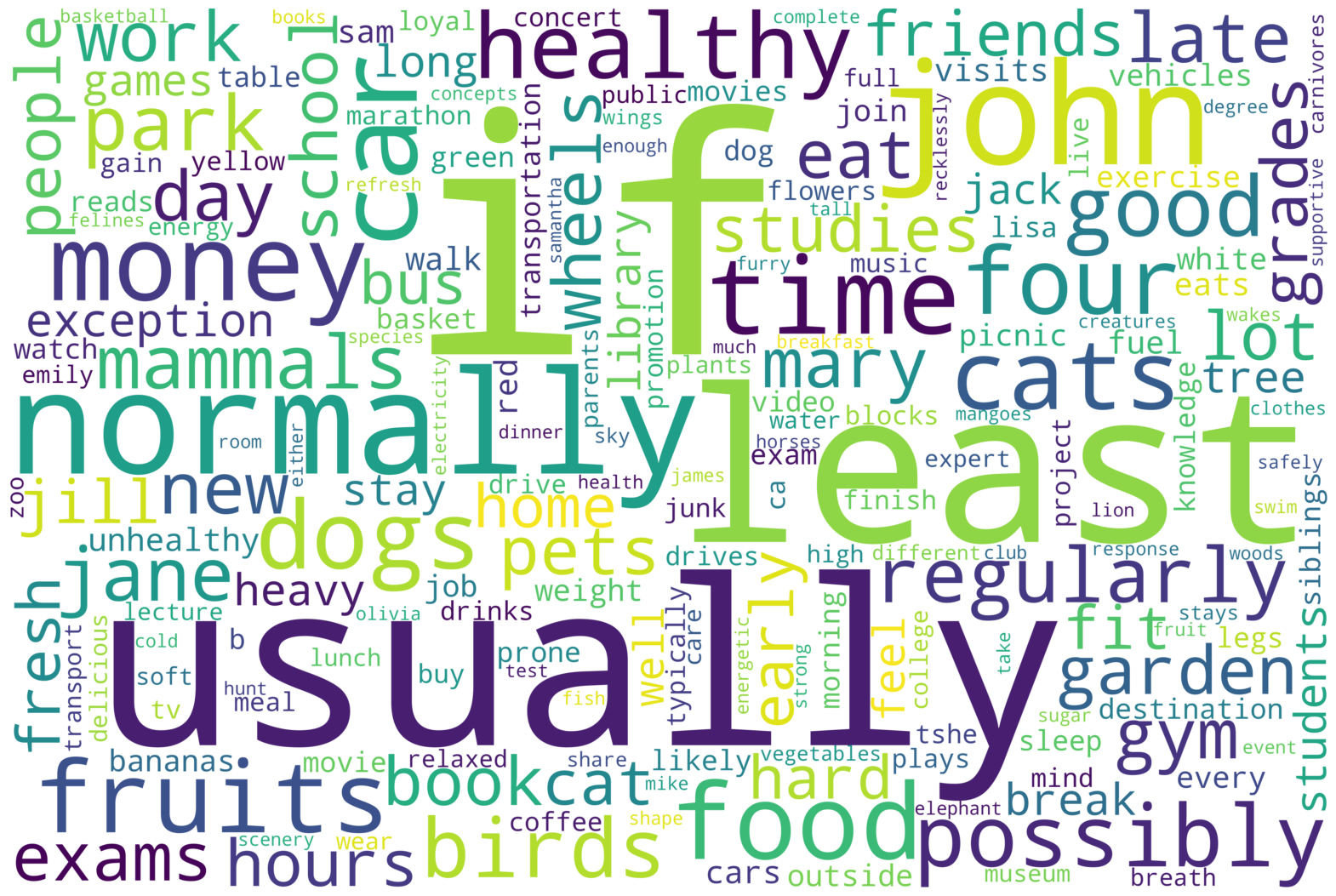}
  \end{center}
  \caption{Word cloud of context present in the LB(eval)}
  \label{fig:wordCloud}
\end{figure}

Figure \ref{fig:wordCloud} provides a word cloud derived from the $LogicBench(Eval)$. This word cloud highlights the logical nature and diversity of our evaluation dataset. Words such as `if', `normally', `usually', and `then' are prominently featured, suggesting their frequent use in the dataset, and suggesting the logical nature of the dataset. Moreover, we can also observe several words consisting of different ontologies such as `cat', `car', `garden', and many more, suggesting diversity in the dataset.

\subsection{Propositional Logic (PL)}

Here, we discuss examples of each inference rule present in the PL of the \textit{LogicBench} as shown in Table \ref{tab:appendix_PL_example}. Table \ref{tab:appendix_PL_example} has context related to the inference rule and different variations of the question according to the rule. For instance, the first row of Table \ref{tab:appendix_PL_example} shows the example for inference rule, \textit{Hypothetical Syllogism (HS)}, formally expressed as $((p\to q))\land (q\to r))\vdash (p\to r)$. The context represents the premise, i.e., $((p\to q))\land (q\to r))$, and the first question (Q1) represents the conclusion, i.e., $p \to r$. Hence, Q1 is labeled as "Yes" since it supports the conclusion given the logical context. Furthermore, Q2 to Q4 represent different variations of the question by utilizing the variables $(p, \neg p, r, \neg r)$. For the HS, given the provided context, Q2 to Q4 contain the variations $\neg p \to r$, $p \to \neg r$, and $\neg p \to \neg r$, respectively, and are labeled as "No" since they do not support the conclusion. 

\subsection{First-Order Logic (FOL)} \label{app:FOL_examples}

Here, we discuss examples of each inference rule and two axioms (i.e., Existential Instantiation and Universal Instantiation) present in the FOL from the \textit{LogicBench} as shown in Table \ref{tab:appendix_FOL_example}. \textit{Existential Generalization (EG)}, formally expressed as ${\displaystyle P\left({a}\right) \Rightarrow \exists xP\left({x}\right)}$ indicates that there is an element $a$ in the domain for which $P(a)$ is true, then we know that $\exists xP\left({x}\right)$ is true. \textit{Universal Instantiation} formally expressed as $\displaystyle \forall x\,A\Rightarrow A\{x\mapsto a\}$ indicates that a statement holds true for all instances (x) within a specific category $A$, hence it is also true for specific instance $a$.

Table \ref{tab:appendix_FOL_example} represents context related to the inference rule and variations of the question. The process of generating data instances for FOL follows a similar approach to that of PL. For example, the first row of Table \ref{tab:appendix_FOL_example} shows the example for axiom, \textit{Existential Instantiation (EI)}, formally expressed as ${\displaystyle \exists xP\left({x}\right) \Rightarrow P\left({a}\right)}$. The context represents the initial premise ${\displaystyle \exists xP\left({x}\right)}$ and the first question (Q1) represents the conclusion, i.e., $P(a)$. Hence, Q1 is labeled as "Yes" since it supports the conclusion given the logical context. Furthermore, we generate the only variant of the question based on $\neg P(a)$ and labeled it as $No$ since it does not support the conclusion. 

\subsection{Non-Monotonic (NM) Reasoning} \label{app:NM_examples}

Here, we discuss examples of each reasoning pattern present in the NM reasoning from the \textit{LogicBench} as shown in Table \ref{tab:appendix_NM_example}. Table \ref{tab:appendix_NM_example} has context related to the reasoning pattern and different variants of the question. For example, the first row of Table \ref{tab:appendix_NM_example} shows the example for \textit{Default Reasoning with Irrelevant Information (DRI)}. For this reasoning, based on the given context, there are also two possible variations of the question where one with a correct conclusion labeled as $Yes$ and another with an incorrect conclusion labeled as $No$. 

\begin{table*}[htbp]
  \centering
    \resizebox{0.9\linewidth}{!}{
    \begin{tabular}{c|m{9cm}|m{10cm}}
    \toprule
    Rule & Context & Question \\ \midrule
    
    HS &  
    If Jim cleans his room, he will receive a reward and if he receives a reward, he will use it to buy a new toy. So, Jim decided to tidy up his room, hoping to earn a reward. He diligently gathered his clothes, organized his toys, and dusted every surface. After a few hours of hard work, Jim's room was spotless.
    & \textbf{Q1:} If Jim cleaned his room, does this imply that he will buy a new toy? (Yes) \newline \textbf{Q2:} If Jim didn't clean his room, does this entail that he won't buy a new toy? (No) \newline \textbf{Q3:} If Jim cleaned his room, does this imply that he won't buy a new toy? (No) \newline \textbf{Q4:} If Jim didn't clean his room, does this imply that he will buy a new toy? (No) \\ \midrule
    
    DS &   
    Either Chloe is studying for her exams, or Mila is going on vacation, or both scenarios are unfolding. It is unclear which of the two options is true at this point. However, one thing is certain - Chloe is not studying for her exams.
    & \textbf{Q1:} Does this entail that Mila is going on vacation? (Yes)\newline \textbf{Q2:} Does this mean that Mila isn't going on vacation? (No) \\ \midrule
    
    CD &   
    If I decide to go for a walk, I will be able to breathe in some fresh air and revitalize myself. On the other hand, if I choose to stay home, I will have the opportunity to enjoy a movie. One thing is certain, either I go for a walk or I stay home. It remains uncertain which of the two options I will ultimately choose. It is entirely possible that I might opt for the walk, or perhaps I will find myself drawn to staying home, or even both possibilities might come to fruition.
    & Can we say at least one of the following must always be true? \newline \textbf{Q1:} (a) I will get some fresh air or (b) I will watch a movie (Yes)\newline \textbf{Q2:} (a) I won't get some fresh air and (b) I will watch a movie (No)\newline \textbf{Q3:} (a) I will get some fresh air and (b) I won't watch a movie (No) \newline \textbf{Q4:} (a) I won't get some fresh air and (b) I won't watch a movie (No)  \\ \midrule
    
    DD &  
    If I decide to order takeout, it means I will save time. On the other hand, if I choose to cook a meal, it means I will save money. The interesting thing is that I am in a situation where I won't be able to save time or money. It is uncertain whether I won't save time or I won't save money, or it could even be both scenarios. The only thing that is clear is that at least one of these possibilities is true.
    & Can we say at least one of the following must always be true?\newline \textbf{Q1:}  (a) I don't order takeout or (b) I don't cook a meal (Yes) \newline \textbf{Q2:} (a) I order takeout and (b) I cook a meal (No) \newline \textbf{Q3:} (a) I don't order takeout and (b) I cook a meal (No) \newline \textbf{Q4:} (a) I order takeout and (b) I don't cook a meal (No)  \\ \midrule
    
    BD &  
    If it is sunny outside, then I will go for a walk. However, if it rains, I will stay inside. Currently, it is uncertain whether it is raining or not, but I do know that at least one of the following is true: either it is raining or I will not go for a walk. It is possible that only one of these statements is true, or perhaps both are true.
    & Can we say at least one of the following must always be true? \newline \textbf{Q1:} (a) we will stay inside  or (b) it is not sunny (Yes) \newline \textbf{Q2:} (a) we will not stay inside and (b) it is sunny (No) \newline \textbf{Q3:} (a) we will stay inside  and (b) it is sunny (No) \newline \textbf{Q4:} (a) we will not stay inside and (b)it is not sunny (No) \\ \midrule
    
    MT &  
    If Mason decides to leave his job, he will not receive any salary. However, against all odds, Mason still receives his salary. He finds himself receiving his regular paycheck.
    & \textbf{Q1:} Does this infer that Mason didn't leave his job? (Yes) \newline \textbf{Q2:} Does this infer that Mason left his job? (No) \\ \midrule
    
    MI &   
    Rohan woke up in the morning and realized that he had forgotten his lunch. Knowing that if he forgets his lunch, he will not eat at school, he felt disappointed.
    & Based on context, can we say, at least one of the following must always be true? \newline \textbf{Q1:} (a) Rohan didn't forget his lunch and (b) he will not eat at school (Yes) \newline \textbf{Q2:} (a) Rohan forgot his lunch and (b) he will eat at school (No)\newline \textbf{Q3:} (a) Rohan forgot his lunch and (b) he will not eat at school (No) \newline \textbf{Q4:} (a) Rohan didn't forget his lunch and (b) he will eat at school (No)  \\ \midrule
    
    CT &   
    At least one of two things is true about Tom - he is either an avid reader or he devours books of all genres. We are unsure which one of these statements is true or if both are true. It could be that only the first statement is true, or only the second statement is true, or even that both are true.
    & Can we say at least one of the following must always be true? \newline \textbf{Q1:} (a) he devours books of all genres or (b) Tom is an avid reader (Yes) \newline \textbf{Q2:} (a) he doesn't devour books of all genres and (b) Tom is an avid reader (No) \newline \textbf{Q3:} (a) he devours books of all genres and (b) Tom isn't an avid reader (No) \newline \textbf{Q4:} (a) he doesn't devour books of all genres and (b) Tom isn't an avid reader (No) \\ 
    \bottomrule
    \end{tabular}%
    }
    \vspace{0.2cm}
    \caption{Examples of context and question-answer pairs for each rule of \textit{Proportional logic} from the LogicBench; HS: Hypothetical Syllogism, DS: Disjunctive Syllogism, CD: Constructive Dilemma, DD: Destructive Dilemma, BD: Bidirectional Dilemma, MT: Modus Tollens, MI: Material Implication, CT: Commutation
    }
    \label{tab:appendix_PL_example}%
\end{table*}%

\begin{table*}[htbp]
  \centering
    \resizebox{0.9\linewidth}{!}{
    \begin{tabular}{c|m{9cm}|m{10cm}}
    \toprule
    Rule & Context & Question \\ \midrule
    
    UI &  
    All students are required to take an examination in order to fulfill the requirements for their degree. Reema, being a student, is also expected to fulfill the requirements.
    & \textbf{Q1:} Does Reema need to take an exam to complete her degree? (Yes) \newline \textbf{Q2:} Does Reema need not to take an exam to complete her degree? (No)  \\ \midrule

    EG  &   
    The marathon race was won by James, who emerged as the champion.
    & \textbf{Q1:} Does this imply that someone won the marathon race? (Yes) \newline \textbf{Q2:} Does this mean that no one won the marathon race? (No) \\ \midrule
    
    MP & 
    If someone is extremely tired, then they will seek some rest and relaxation. Today, Jack finds himself utterly exhausted.
    & \textbf{Q1:} Does this entail that he will take rest? (Yes) \newline \textbf{Q2:} Does this entail that he won't take rest? (No)  \\ \midrule
    
    HS & 
    If all the necessary supplies have been purchased by someone, then they can initiate the project. Once the project is started by someone, they will complete it within the expected time-frame.
    & \textbf{Q1:} If Lily bought all the necessary supplies, does this mean that she will finish it on time? (Yes) \newline \textbf{Q2:} If Lily didn't buy all the necessary supplies, does this imply that she won't finish it on time? (No) \newline \textbf{Q3:} If Lily bought all the necessary supplies, does this entail that she won't finish it on time? (No) \newline \textbf{Q4:} If Lily didn't buy all the necessary supplies, does this imply that she will finish it on time? (No) \\ \midrule
    
    DS & 
    It is known that one of the following options is true: someone goes to a museum or someone visits a park. The specific scenario could involve only the option to go to a museum being true, or only the option to visit a park being true, or both options being true. However, it is stated that Jill is unable to go to a museum.
    & \textbf{Q1:} Does this imply that she can visit a park? (Yes) \newline \textbf{Q2:} Does this entail that she can't visit a park? (No) \\ \midrule
    
    CD &  
    If someone is painting a picture, then they will frame it. similarly, the natural course of action for a writer would be to publish their completed story. in this scenario, it is certain that at least one of the following statements holds true: (1) john is currently engrossed in painting a picture, or (2) john is currently immersed in the act of writing a story. it should be emphasized that we are unaware of which statement specifically applies, as there is a possibility that either (1) alone is true, or (2) alone is true, or even that both (1) and (2) are simultaneously true.
    & Can we say at least one of the following must always be true? \newline \textbf{Q1:} (a) he will frame it and (b) he will publish it. (Yes) \newline \textbf{Q2:} (a) he won't frame it and (b) he will publish it. (No) \newline \textbf{Q3:} (a) he will frame it and (b) he won't publish it. (No) \newline \textbf{Q4:} (a) he won't frame it and (b) he won't publish it. (No)   \\ \midrule
    
    DD & 
    If someone is taking care of their health, then they will be fit. However, indulging in unhealthy habits can make individuals susceptible to various diseases. The truth is, we can be certain about at least one of the following possibilities: either Jenny won't be fit and healthy, or she won't be prone to diseases. It is important to note that we are unaware of which statement is accurate. It could be the case that only the first statement is true, only the second statement is true, or both statements hold validity.
    & Can we say at least one of the following must always be true? \newline \textbf{Q1:} (a) Jenny doesn't take care of her health and (b) she doesn't indulge in unhealthy habits (Yes) \newline \textbf{Q2:} (a) Jenny takes care of her health and (b) she indulges in unhealthy habits (No) \newline \textbf{Q3:} (a) Jenny doesn't take care of her health and (b) she indulges in unhealthy habits (No) \newline \textbf{Q4:} (a) Jenny takes care of her health and (b) she doesn't indulge in unhealthy habits (No) \\ \midrule
    
    BD &  
    If an individual consumes a significant amount of water, they will experience a state of hydration. Conversely, if excessive amounts of sugar are ingested, a sugar crash will ensue. It is known that at least one of the following statements is true: either the Jane consumes ample water or she will not experience a sugar crash. However, the actual veracity of either statement remains ambiguous, as it could be the case that only the first statement is true, only the second statement is true, or both statements are true.
    & Can we say at least one of the following must always be true?  \newline \textbf{Q1:} (a) she will feel hydrated and (b) she doesn't eat too much sugar (Yes) \newline \textbf{Q2:} (a) she won't feel hydrated and (b) she eats too much sugar (No) \newline \textbf{Q3:} (a) she will feel hydrated and (b) she eats too much sugar (No) \newline \textbf{Q4:} (a) she won't feel hydrated and (b) she doesn't eat too much sugar (No) \\ \midrule
    
    MT &  
    If someone decides to go to the park, it is required that they wear a mask. However, in this particular situation, John does not wear a face covering.
    & \textbf{Q1:} Does this imply that John doesn't visit the park? (Yes) \newline \textbf{Q2:} Does this entail that John visits the park? (No) \\ \bottomrule
    \end{tabular}%
    }
    \vspace{0.2cm}
    \caption{Examples of context and question-answer pairs for each rule of \textit{First order logic} from the LogicBench.}
  \label{tab:appendix_FOL_example}%
\end{table*}%

\begin{table*}[htbp]
  \centering
  \resizebox{\linewidth}{!}{
    \begin{tabular}{c|m{9cm}|m{10cm}} \toprule
    Rule & Context & \multicolumn{1}{l}{Question} \\ \midrule
    
    DRI & 
    Once upon a time, in a land filled with animals, there were two popular mammalian creatures, cats and dogs. Mammals typically possessed a coat of fur, which kept them warm and protected. However, cats were an exception to this rule, as their bodies lacked fur. Nonetheless, both cats and dogs were beloved by many for their unique traits. Dogs, known for their loyalty, were particularly cherished by humans.
    & \textbf{Q1:} Does this imply that dogs have fur? (Yes) \newline \textbf{Q2:} Does this entail that dogs don't have fur? (No) \\ \midrule
    
    DRS & 
    John and Mary were expecting their first child, filled with the anticipation and excitement that all parents feel. Parents are usually loving and supportive. Parents are normally responsible. However, something seemed amiss in their relationship. Mary, usually affectionate and caring, seemed distant and uninvolved. On the other hand, John, known for his responsible nature, started neglecting his duties and became unreliable.
    & \textbf{Q1:} Does this imply that Mary is responsible and John is loving and supportive? (Yes)\newline \textbf{Q2:} Does this entail that Mary isn't responsible and John is loving and supportive? (No)\newline \textbf{Q3:} Does this imply that Mary is responsible and John isn't loving and supportive? (No)\newline \textbf{Q4:} Does this entail that Mary isn't responsible and John isn't loving and supportive? (No) \\ \midrule
    
    DRD &  
    Jenny and Anna are known for their tall stature, which is often associated with playing basketball. However, Anna might be an exception to this norm.
    & \textbf{Q1:} Does this entail that Jenny plays basketball? (Yes)\newline \textbf{Q2:} Does this mean that Jenny doesn't play basketball? (No) \\ \midrule 
    
    DRO &  
    In the bird kingdom, there are many different species that possess unique characteristics. One such species is the hummingbird, known for its ability to hover in mid-air and its vibrant colors. While most birds engage in the annual migration south for the winter, the hummingbird chooses to stay put and brave the cold weather. This decision sets the hummingbird apart from its fellow avian companions, as it relies on its resilience and resourcefulness to survive the harsh conditions.
    &  \textbf{Q1:} Does this mean that all other birds than hummingbirds migrate south for the winter? (Yes)\newline \textbf{Q2:} Does this mean that all other birds than hummingbirds don't migrate south for the winter? (No) \\ \midrule
    
    RE1 & 
    In a world where animals are often regarded as intelligent creatures, there is a captivating tale that revolves around cats, dogs, and horses. It is commonly believed that most animals possess a level of intellect. However, there is an intriguing twist to this belief as it is known that either cats or dogs are not considered particularly intelligent. As the story unfolds, we delve into the lives of these remarkable creatures, their interactions, and the unique qualities that each of them possesses.
    & \textbf{Q1:} Does this entail that horses are considered to be intelligent creatures and exactly one of the cats or dogs is not considered intelligent? (Yes)\newline \textbf{Q2:} Does this mean that horses aren't considered to be intelligent creatures and exactly one of cats or dogs is not considered intelligent? (No)\newline \textbf{Q3:} Does this mean that horses are considered to be intelligent creatures and exactly one of cats or dogs is considered intelligent? (No)\newline \textbf{Q4:} Does this implies that horses aren't considered to be intelligent creatures and exactly one of cats or dogs is considered intelligent? (No) \\ \midrule
    
    RE2 &  
    In the realm of cat communication, meowing serves as a fundamental aspect of their vocal repertoire. However, intriguingly enough, there exists a distinct species of cat that deviates from this conventional norm. This peculiar feline defies the expectations associated with its kind by refraining from emitting any meows whatsoever.
    & \textbf{Q1:} Does this entail that exactly one species of cat doesn't meow? (Yes)\newline \textbf{Q2:} Does this imply that exactly one species of cat meows? (No) \\ \midrule
    
    RE3 &   
    In a world where cars were known for having four wheels, it was considered a common fact that wheels typically came equipped with spokes. However, amidst this widespread understanding, there was an exception. At least one wheel defied this norm and stood out from the rest by not having any spokes at all.
    &  \textbf{Q1:} Does this imply that cars have four wheels with spokes? (Yes)\newline \textbf{Q2:} Does this mean that cars don't have four wheels with spokes? (No) \\ \midrule
    
    RAP & 
    In the midst of a heated argument, John adamantly claims that sally was present at the store. However, Jane strongly opposes John's assertion, insisting that Sally was indeed absent from the store.
    &  \textbf{Q1:} If John's evidence is more reliable than Jane's, does this mean that Sally was in the store? (Yes)\newline \textbf{Q2:} If John's evidence is more reliable than Jane's, does this mean that Sally wasn't in the store? (No)\newline \textbf{Q3:} If John's evidence is less reliable than Jane's, does this entail that Sally was in the store? (No)\newline \textbf{Q4:} If John's evidence is less reliable than Jane's, does this imply that Sally wasn't in the store? (Yes) \\ \bottomrule
    \end{tabular}%
    }
    \vspace{0.2cm}
    \caption{Examples of context and question-answer pairs for each rule of \textit{Non-monotonic logic} from the LogicBench.}
  \label{tab:appendix_NM_example}%
\end{table*}%

\section{Examples of NL Conversion} \label{app:NL_conversion}
This section illustrates the way natural language logical context and questions are created using the generated sentence in Stage 1 in addition to prompt-based templatized-context to narrative conversion.
Table \ref{tab:NL_conversion_examples} shows examples of how context and question are generated from sentences corresponding to each inference rule for PL and FOL. Similarly, Table \ref{tab:NL_conversion_examples_2} shows examples of NM reasoning. From Table \ref{tab:NL_conversion_examples}, we can see an example of sentence pairs $(p,q)$ and their corresponding negation pairs $(\neg p, \neg q)$ for the `modus tollens' inference rule from PL. These pairs are utilized to generate templatized logical context which is converted to narrative using a specific prompt[1\ref{PBC:PL_prompt}] designed for PL. In addition to the narrative, the question and its variations are also created using these sentence pairs. Similarly, in the second row, we have four generic rules with variable $x$ $(p(x), q(x), r(x), s(x))$ and their specific cases (i.e., $x=a$), along with their respective negative sentence pairs [$(p(a), \neg p(a))$, $(q(a), \neg q(a))$, $(r(a), \neg r(a))$, $(s(a), \neg s(a))$]. In the same manner, as PL, these examples demonstrate the generation of narrative and questions for the FOL inference rule called `Bidirectional Dilemma (BD)', as shown in Table \ref{tab:NL_conversion_examples} using specific prompt[2\ref{PBC:FOL_prompt}] designed for FOL. From Table \ref{tab:NL_conversion_examples_2}, the first row presents an example of narrative and questions generated from a sentence pair for the `Default Reasoning with Irrelevant Information (DRI)' from NM reasoning. In this specific instance, the generated sentences are $(p, q, r, s, t)$, and the negation is only required for the sentence $t$. Therefore, there is a single negation pair $(t, \neg t)$, which is used to generate questions specific to the `DRI'. Same as PL and FOL, templatized context is also converted to narrative using a prompt[3\ref{PBC:NM_prompt}] created for NM. Each prompt is shown below for PL, FOL, and NM templatized context to narrative conversion.


\textbf{1. Prompt for templatized to narrative conversion for PL(MT):}
\begin{tcolorbox} \label{PBC:PL_prompt}
\textbf{Rule:} \textbf{Condition:} If p then q; \textbf{Situation:} not q \newline
\textbf{Context:} Templatized context\newline
Improve the context in human-like language and make story with rephrased sentences.\newline 
\textbf{Instructions for generating a good story:}\newline 
1. While generating a story, Use the rephrased sentences from the context of the story.\newline 
2. Make sure to include sentences corresponding to the condition and situation from the rule in the story.\newline 
3. Do not add any other extra information.\newline 
4. To generate a story, DO NOT change the name of the main character from the context, if there's any. If the main character within the context is 'I', please generate a story with 'I' as the main character.\newline 
5. Generate only one paragraph with rephrased sentences. 
\end{tcolorbox}

\textbf{2. Prompt for templatized to narrative conversion for FOL:}
\begin{tcolorbox} \label{PBC:FOL_prompt}
\textbf{Context: } Templatized context\newline
Improve the context in human-like language and make a rephrase sentences around it.\newline 
\textbf{Instructions for generating a good story:}\newline
1. Make sure to include only rephrased sentences in the story.\newline
2. Do not add any other extra information and do not reach any conclusion based on context.\newline
3. To generate a story, use only generic sentences without mentioning any main character.\newline
4. Extract a specific case sentence from the context, rephrase it, and append it at the end of the context only.\newline
5. Generate only one paragraph with rephrased sentences.\newline
6. DO NOT add any prefixes.
\end{tcolorbox}

\textbf{3. Prompt for templatized to narrative conversion for NM:}
\begin{tcolorbox} \label{PBC:NM_prompt}
\textbf{Context: } Templatized context\newline
Improve the context in human-like language and rephrase sentences around it.\newline 
\textbf{Instructions for generating a good story:}\newline
1. Make sure to include only rephrased sentences in the story.\newline
2. Do not add any other extra information, but add some story-like context.\newline
3. To generate a story, DO NOT change the name of the main character from the context, if there's any. Use the same names/words for main character. If the main character within the context is 'I', please generate a story with 'I' as the main character.\newline
4. Generate only one paragraph with rephrased sentences.
\end{tcolorbox}

To ensure the quality of the narrative in the \textit{LogicBench(Eval)} for task 1 and task 2, we have created category-specific prompts to convert the templatized context to convert more human-like narrative. In total, we have created three different prompts each for PL[1\ref{PBC:PL_prompt}], FOL[2\ref{PBC:FOL_prompt}], and NM[3\ref{PBC:NM_prompt}. The prompts are designed to ensure that the logical connection is established in the narrative. Each rule of PL, FOL, and NM has a unique logical progression that should be followed in the narrative is the motivation for us to go with three different instruction-based prompts. In the prompt, used for PL rules, the logical rule is also mentioned in two parts "Condition" and "Situation". Consider the example of logical rule MT - "If p then q; not q; therefore not p", the templatized context will have "If p then q; not q;" so condition will be the first part "If p then q" while the situation will be the second part "not q". In other terms, the specific conditions are the rules to be followed while the situation is the case present in the logical rule. In contrast, for FOL, we do not need to focus more on the specific condition rather we have to make sure a generalized case is present in the narrative with 1-2 specific sentences related to the rule. For NM, we do not have such a rule instead we have a logical connection between sentences and hence we go with the instruction-based prompting.

\section{Task Instance Generation}
This section discusses the task instance generation step, which is the last step in Fig. \ref{fig:approach} in detail. The overall section is comprised of two subsections for BQA and MCQA. The subsection related to BQA discusses the variation generation step in detail with question generation while the MCQA subsection provides details of prompts used for incorrect option generation along with five pre-defined sets of questions. 

\subsection{Task 1- BQA (variation generation)}
As seen in Fig. \ref{fig:approach}, BQA has a narrative along with question, and answer pairs with variation in question and corresponding answer. As discussed in the main paper Data creation\ref{sub:3.2}, the narrative is created using the pre-defined rules for PL, FOL, and NM while a question is asked based on what can be entailed from the given context. For example, HS can be defined as "If p then q; if q then r; therefore, if p then r" and the narrative will have "If p then q; if q then r" and the question is asked as "Can we conclude if p then r?" having answer as "yes". Now, other variations in the question are asked in the following ways by negating the sentence (p and r) comb: Variation 1: "Can we conclude if $\neg$p then r?", Variation 2: "Can we conclude if p then $\neg$r?", Variation 3: "Can we conclude if $\neg$p then $\neg$r?". If there's only one axiom (p) present in the question then there are only a 2 variations that can be made asking about (p) and ($\neg$p). 

\subsection{Task 2-MCQA (Question selection and incorrect option generation)} \label{E.2} 
We have MCQA as task 2 in LogicBench. In this, we have one correct option from the four options, and three options are incorrectly generated using prompting. The question is a bit different from the BQA question formation as we have MCQA and have to identify which conclusion can be derived from the narrative. The question is randomly selected from the pre-defined set of questions. Here, the five different questions are as follows: 
\begin{enumerate}
    \item What would be the most appropriate conclusion based on the given context?
	\item Considering the provided context, what conclusion would be deemed most suitable?
	\item In light of the context provided, what conclusion can be considered the most appropriate?
	\item Based on the context, what conclusion would be deemed most suitable?
	\item Taking into account the context provided, what conclusion would be most appropriate?
\end{enumerate}
The narrative for MCQA is the same as the BQA's narrative and there's no change. Comparing BQA methodology with MCQA, MCQA's correct option is the question asked in the BQA which can be concluded from the logical rule present in the narrative. For example, HS can be defined as "If p then q; if q then r; therefore, if p then r" and the narrative will have "If p then q; if q then r" and the question is randomly chosen from the set of questions as mentioned above and the correct option from the different multiple choice will be "if p then r". Based on the information present, we have generated an incorrect option using prompting[4\ref{PBC:MCQ}]. Refer this link for more examples - \url{https://anonymous.4open.science/r/LogicBench-EEBB}.
    
\textbf{4. Prompt for incorrect option generation:} 
\begin{tcolorbox} \label{PBC:MCQ}
\textbf{Context:} Narrative\newline
\textbf{Question:} Randomly chosen question\newline
\textbf{Correct option:} Conclusion from narrative \newline\newline
Create three more irrelevant, incorrect options to create multiple choice. Sometime use different persons name as well in creating incorrect options.\newline
\textbf{Instruction to generate options:}\newline
1. Generate only three unique choices each in new line. \newline
2. Do not use any prefix while generating.
\end{tcolorbox}

\section{Mitigation of Errors in \textit{LogicBench}} \label{app:error_mitigation}
While validating \textit{LogicBench(Eval)}, we encountered errors within the synthetically generated narrative-based context. We mitigate these errors manually, categorizing them into two groups: (i) eliminating leakage of logical conclusions and (ii) ensuring the inclusion of intended logical premises. In the first category, we found $\sim15\%$ of the narrative-based contexts (out of 500 total instances) were found to explicitly present the logical conclusion as a response to the question, bypassing the logical derivation process. This enables the model to extract the final logical conclusion from the context rather than derive it logically. In the second category, we found $\sim8\%$ of the narrative-based contexts (out of 500 total instances) where narration lacked some necessary premise sentences crucial for reaching a logical conclusion. To address this issue, we manually incorporated those sentences to ensure the quality of the data. For the MCQA task, during the generation of three incorrect options, we found instances where the model produces two semantically similar options, resulting in the creation of ambiguous choices in $\sim11\%$ of the cases (out of a total of 500 instances). We manually mitigated all these errors from the data instances ensuring the high quality of our validation data. We believe that these two versions aim to accommodate different evaluations to explore the logical reasoning capabilities of LLMs.

\section{Few-shot Experiments} \label{app:few_shot}

This section discusses the performance of the different LLMs in a few-shot setting on the \textit{LogicBench(Eval)}$_{BQA}$. Here, we only present a case study on the BQA task. For the fair comparison with Table \ref{tab:main_result}, we analyze an average performance across $A(Yes)$. Table \ref{tab:few_shot_result} shows the performance for each inference rule and reasoning patterns achieved by Llama-2, Mistral, Gemini, ChatGPT, and GPT-4. 

As suggested in \citep{lu-etal-2022-fantastically}, prompting models are sensitive to in-context examples. Hence, we see mixed performance in Table \ref{tab:few_shot_result} across all models. From Table \ref{tab:few_shot_result}, we can observe that in-context examples are helpful for Llama-2 since it consistently outperforms zero-shot baselines by large margins in terms of $A(Yes)$. Llama-2 is remarkably good at following the in-context exemplars and mimicking the process to reach the correct conclusions. Thus, leveraging the in-context exemplars, Llama-2 achieves high accuracy in a few-shot setting. Specifically, Mistral consistently shows degraded performance for all logic types. However, ChatGPT improves performance on NM reasoning. Improved performance in NM reasoning demonstrates that the inclusion of in-context examples enhances the ability of these models to comprehend the nuanced meanings of logical terms such as ``usually'' and ``typically''. In particular, we see that Gemini and GPT-4 improve performance on PL and FOL, respectively, but show competitive performance on NM. 

\section{Experimental Setup} \label{app:exp_prompts}
\subsection{Extended Discussion on Experiments}
\paragraph{Zero-shot setting} 
We evaluate GPT-4, and ChatGPT (GPT-3.5-Turbo) by utilizing their APIs provided by OpenAI\footnote{\url{https://platform.openai.com/docs/guides/gpt}}. We evaluate Google Gemini-Pro by utilizing its API provided by Google\footnote{\url{https://ai.google.dev/}} The evaluation is conducted on the versions of GPT-4, ChatGPT, and Gemini released in January 2024. It's important to note that these models are regularly updated, so when reproducing the results presented in Table \ref{tab:main_result} and Table \ref{tab:mcq_result} (main paper), there is a possibility of variations. For Llama-2 and Mistral, we utilize the 7B-Chat, and 7B-Instruct-v0.2 versions, respectively, from the huggingface model repository\footnote{\url{https://huggingface.co/models}}.

\paragraph{Experiments on other logic datasets}
In single and multi-task experiments on other logic datasets, we fine-tune the T5-large model for 10 epochs with a batch size of 16, 1024 maximum input length, an adaptive learning rate of $5e-05$, and an AdamW optimizer for each experiment. All experiments are performed using NVIDIA RTX A6000 GPUs.

\subsection{Prompts} \label{app:three_prompts}
All the experiments conducted in the zero-shot setting were performed using three distinct prompts. The reported results in Table \ref{tab:main_result} (main paper) represent the average performance across these prompts. All the prompts follow the common pattern which includes task description and formatting instructions. The following are the three different prompts utilized in the experiments:

\textbf{Prompt 1:}
\begin{tcolorbox}
Given the context and question, think step-by-step logically to answer the question. Answer the question ONLY in `yes' or `no'. Please use the below format:
\newline \textbf{Context}: [text with logical rules] \newline \textbf{Question}: [question based on context] \newline \textbf{Reasoning steps}: [generate step-by-step reasoning] \newline \textbf{Answer}: Yes/No
\end{tcolorbox}

\textbf{Prompt 2:}
\begin{tcolorbox}
Let's think step-by-step to answer the question given context. Answer the question ONLY in `yes' or `no'. Please use the below format:
\newline \textbf{Context}: [text with logical rules] \newline \textbf{Question}: [question based on context] \newline \textbf{Reasoning steps}: [generate step-by-step reasoning] \newline \textbf{Answer}: Yes/No
\end{tcolorbox}

\textbf{Prompt 3:}
\begin{tcolorbox}
Given the context that contains rules of logical reasoning in natural language and question, perform step-by-step reasoning to answer the question. Based on context and reasoning steps answer the question ONLY in `yes' or `no'. Please use the below format:
\newline \textbf{Context:} [text with logical rules] \newline \textbf{Question:} [question based on context] \newline \textbf{Reasoning steps:} [generate step-by-step reasoning] \newline \textbf{Answer:} Yes/No
\end{tcolorbox}

For MCQA task, the reported results in Table \ref{tab:mcq_result} (main paper) represent the average performance across three prompts. All the prompts follow the common pattern which includes task description and formatting instructions. The following are the three different prompts utilized in the experiments:

\textbf{Prompt 1:}
\begin{tcolorbox}
Given the context that contains rules of logical reasoning in natural language, question, and options, perform step-by-step reasoning to answer the question. Answer the question ONLY in 'choice\_1' or 'choice\_2' or 'choice\_3' or 'choice\_4'. DO NOT generate any text with it. Based on context and reasoning steps, choose the correct option. \newline
Please use the below format: \newline
\textbf{Context:} [text with logical rules] \newline
\textbf{Question:} [question based on context] \newline
\textbf{Options:} [four different choices] \newline
\textbf{Reasoning steps:} [generate step-by-step reasoning] \newline
\textbf{Answer:} choice\_1 / choice\_2 / choice\_3 / choice\_4
\end{tcolorbox}

\textbf{Prompt 2:}
\begin{tcolorbox}
Let's think step-by-step to choose the correct option for the question given context. Answer the question ONLY in 'choice\_1' or 'choice\_2' or 'choice\_3' or 'choice\_4'. DO NOT generate any text with it. Based on context and reasoning steps, choose the correct option. \newline
Please use the below format: \newline
\end{tcolorbox}
\begin{tcolorbox}
\textbf{Context:} [text with logical rules] \newline
\textbf{Question:} [question based on context] \newline
\textbf{Options:} [four different choices] \newline
\textbf{Reasoning steps:} [generate step-by-step reasoning] \newline
\textbf{Answer:} choice\_1 / choice\_2 / choice\_3 / choice\_4
\end{tcolorbox}

\textbf{Prompt 3:}
\begin{tcolorbox}
Given the context, question, and options, think step-by-step logically to answer the question by choosing the correct option. Answer the question ONLY in 'choice\_1' or 'choice\_2' or 'choice\_3' or 'choice\_4'. DO NOT generate any text with it. Based on context and reasoning steps, choose the correct option. \newline
Please use the below format: \newline
\textbf{Context:} [text with logical rules] \newline
\textbf{Question:} [question based on context] \newline
\textbf{Options:} [four different choices] \newline
\textbf{Reasoning steps:} [generate step-by-step reasoning] \newline
\textbf{Answer:} choice\_1 / choice\_2 / choice\_3 / choice\_4
\end{tcolorbox}

\section{Further Discussion on Results}\label{app:discussion_results}

\paragraph{Effect on other logic datasets} 
Our experiments were carried out in two settings: single-task (fine-tuning and evaluation on one dataset) and multi-task (fine-tuning on all four datasets combined, with separate evaluations for each dataset). App. \ref{app:exp_prompts} describes a detailed experimental setup.
Table \ref{tab:other_results} represents the accuracy comparison between LogicT5 and baseline T5-large in both single-task and multi-task settings.


The results indicate that training LLMs on \textit{LogicBench(Aug)} has a greater impact on logic datasets that primarily focus on logical reasoning, such as FOLIO and LogicNLI. Hence, we can observe that LogicT5 consistently outperforms the baseline for LogicT5 and FOLIO. However, LogiQA and ReClor encompass other forms of reasoning in addition to logical reasoning, hence, LogicT5 demonstrates competitive performance on them.

\begin{table}[ht]
\centering
\small
\resizebox{\linewidth}{!}{
\begin{tabular}{cccccc}
\toprule
Methods                      & Models   & LogiQA        & FOLIO         & LogicNLI      & ReClor        \\ \midrule
\multirow{2}{*}{Single-Task} & T5-large & 16.8          & 69.6          & 82.3          & 35.4 \\ 
                             & LogicT5  & \textbf{16.9} & \textbf{71.2} & \textbf{84.4} & \textbf{36.8}          \\ \midrule \midrule
\multirow{2}{*}{Multi-Task}  & T5-large & \textbf{21.8} & 83.8          & 68.2          & \textbf{42.8} \\
                             & LogicT5  & 19.7          & \textbf{85.6} & \textbf{69.8} & 40.0          \\ \bottomrule
\end{tabular}
}
\caption{Performance comparison between LogicT5 and baseline T5-large in terms of accuracy.}
\label{tab:other_results}
\end{table} 
\begin{table*}[ht]
\centering
\resizebox{0.73\linewidth}{!}{
\begin{tabular}{m{0.8cm}|m{7cm}|m{12cm}}
\toprule
Rule    &  \multicolumn{1}{c|}{Generate Sentences in Step 1}  & \multicolumn{1}{c}{NL logical expressions} \\ \midrule

MT & p: Liam finished his work early. \newline $\sim$p: Liam did not finish his work early. \newline q: He will order pizza for dinner. \newline $\sim$q: He will not order pizza for dinner. & \textbf{Context:} Liam had finished his work early for the day, which meant that he would typically have ordered pizza for dinner. However, on this particular day, he decided against ordering pizza and opted for something else instead. \newline \newline \textbf{Question:} Does this imply that liam didn't finish his work early? \\ \midrule

BD & p(x): someone drinks lots of water \newline q(x): they will feel hydrated \newline r(x): they eat too much sugar \newline s(x): they will experience a sugar crash \newline p(a): Jane drinks lots of water \newline $\sim$p(a): Jane does not drink lots of water \newline q(a): she will feel hydrated \newline $\sim$q(a): she will not feel hydrated \newline r(a): she eats too much sugar \newline $\sim$r(a): she does not eat too much sugar \newline s(a): she will experience a sugar crash \newline $\sim$s(a): she will not experience a sugar crash & \textbf{Context:} If an individual consumes a significant amount of water, they will experience a state of hydration. Conversely, if excessive amounts of sugar are ingested, a sugar crash will ensue. It is known that at least one of the following statements is true: either the Jane consumes ample water or she will not experience a sugar crash. However, the actual veracity of either statement remains ambiguous, as it could be the case that only the first statement is true, only the second statement is true, or both statements are true. \newline \newline \textbf{Question:} Can we say at least one of the following must always be true? (a) she will feel hydrated and (b) she doesn't eat too much sugar \\ \midrule

MP & 
p(x): someone is exhausted. \newline
q(x): they will take a rest. \newline
p(a): Jack is exhausted. \newline
$\sim$p(a): Jack is not exhausted. \newline
q(a): he will take a rest. \newline 
$\sim$q(a): he will not take a rest. &
\textbf{Context:}  If someone is extremely tired, then they will seek some rest and relaxation. Today, Jack finds himself utterly exhausted. \newline \newline \textbf{Question:} Does this entail that he will take rest? \\ \midrule
    
    
DS & 
p: Levi is not studying for his exams \newline
$\sim$p: Levi is studying for his exams \newline
q: Maya is writing a book \newline
$\sim$q: Maya is not writing a book &
\textbf{Context:} Either Levi is not studying for his exams or Maya is writing a book, or maybe both. It was unclear which choice to make as he didn't know if either of the options was true. However, it turned out that Levi decided to prioritize his exams and focus on studying. He knew that in order to succeed, he needed to dedicate his time and energy to preparing for the upcoming tests. \newline \newline
\textbf{Question:} Does this mean that Maya is writing a book?\\ \midrule

HS & 
p(x): someone buys all the necessary supplies. \newline
q(x): they can start the project. \newline
r(x): they will finish it on time. \newline
p(a): Lily bought all the necessary supplies.\newline 
$\sim$p(a): Lily did not buy all the necessary supplies. \newline
q(a): She can start the project.\newline
$\sim$q(a): She can not start the project.\newline
s(a): She will finish it on time. \newline
$\sim$s(a): She will not finish it on time.&
\textbf{Context:} If all the necessary supplies have been purchased by someone, then they can initiate the project. Once the project is started by someone, they will complete it within the expected timeframe. \newline \newline \textbf{Question:} If lily bought all the necessary supplies, does this mean that she will finish it on time? \\ \midrule
    
CD &  
p: Harry goes to the park. \newline
$\sim$p: Harry does not go to the park. \newline
q: he will have a picnic with his family \newline
$\sim$q: he will not have a picnic with his family \newline
r: he goes to the beach \newline
$\sim$r: he does not go to the beach \newline
s: he will swim in the ocean \newline
$\sim$s: he will not swim in the ocean
& 
\textbf{Context:} Harry had two options for his day off: going to the park or going to the beach. If he decided to go to the park, it meant that he would have a lovely picnic with his family. On the other hand, if he chose the beach, he would be able to enjoy a refreshing swim in the ocean. The only thing that was certain was that either Harry would go to the park or he would go to the beach. It was unclear which option he would ultimately choose, as it could be that he would go to the park alone, go to the beach alone, or even do both activities. \newline \newline
\textbf{Question:} Can we say at least one of the following must always be true? (a) he will have a picnic with his family and (b) he will swim in the ocean \\ \midrule
    
DD & 
p: I order takeout \newline
$\sim$p: I did not order takeout \newline
q: I will save time \newline
$\sim$q: I will not save time \newline
r: I cook a meal \newline
$\sim$r: I did not cook a meal \newline
s: I will save money \newline
$\sim$s: I will not save money 
&
\textbf{Context:} If I decide to order takeout, it means I will save time. On the other hand, if I choose to cook a meal, it means I will save money. The interesting thing is that I am in a situation where I won't be able to save time or money. It is uncertain whether I won't save time or I won't save money, or it could even be both scenarios. The only thing that is clear is that at least one of these possibilities is true. \newline \newline
\textbf{Question:} Can we say at least one of the following must always be true? (a) i don't order takeout and (b) i don't cook a meal \\ \midrule
    
    
CT &  
p: Tom is an avid reader \newline
$\sim$p: Tom is not an avid reader \newline
q: he devours books of all genres \newline
$\sim$q: he does not devour books of all genres
& 
\textbf{Context:} At least one of two things is true about Tom - he is either an avid reader or he devours books of all genres. We are unsure which one of these statements is true or if both are true. It could be that only the first statement is true, or only the second statement is true, or even that both are true. \newline \newline
\textbf{Question:} Can we say at least one of the following must always be true? (a) he devours books of all genres or (b) tom is an avid reader \\ \midrule
    
MI &  
p: he is not eating healthy \newline
$\sim$p: he is eating healthy \newline
q: he will not gain weight \newline
$\sim$q: he will gain weight & 
\textbf{Context:} Once upon a time, there was a guy named John who was very conscious about his weight. He knew that if he didn't eat healthy, he wouldn't be able to maintain his desired weight. John understood the condition that if he didn't eat healthy, he wouldn't be able to gain weight. \newline \newline 
\textbf{Question:} Based on context, can we say, at least one of the following must always be true? (a) he is eating healthy and (b) he will not gain weight \\ \midrule
    
EG &  
p(x): someone has coding skills \newline
a: Sheila \newline
p(a): Sheila is a proficient programmer
& 
\textbf{Context:} Sheila possesses exceptional coding skills and is highly proficient in programming. She has mastered various programming languages and is known for her ability to write efficient and error-free code. Sheila's expertise in programming allows her to solve complex problems with ease. Her colleagues often seek her assistance when facing technical challenges as she consistently delivers high-quality solutions. \newline \newline
\textbf{Question:} does this mean that someone has coding skills?\\ \midrule

UI &  
p(x): students need to take an exam to complete their degree \newline
a: Reema \newline
p(a): Reema is a student.
&
\textbf{Context:} All students are required to take an examination in order to fulfill the requirements for their degree. Reema, being a student, is also expected to fulfill the requirements. \newline \newline
\textbf{Question:} Does reema need to take an exam to complete her degree?\\ \bottomrule

\end{tabular}
}
\vspace{0.2cm}
\caption{Illustrative examples of NL logical context and questions created using sentences that are generated in \textsection \ref{sec:sent_gen} for rules covered in PL and FOL for BQA.}
\label{tab:NL_conversion_examples}
\end{table*}
\begin{table*}[ht]
\centering
\resizebox{0.75\linewidth}{!}{
\begin{tabular}{m{1cm}|m{7cm}|m{10cm}}
\toprule
Rule    &  \multicolumn{1}{c|}{Generate Sentences in Step 1}  & \multicolumn{1}{c}{NL logical expressions} \\ \midrule

    DRI &  
    p: Cats and dogs are mammals. \newline
    q: Mammals typically have fur. \newline
    r: Cats don't have fur. \newline
    s: Dogs are loyal animals. \newline
    t: Dogs have fur. \newline
    $\sim$t: Dogs don't have fur.
    & \textbf{Context:} Once upon a time, in a land filled with animals, there were two popular mammalian creatures, cats and dogs. Mammals typically possessed a coat of fur, which kept them warm and protected. However, cats were an exception to this rule, as their bodies lacked fur. Nonetheless, both cats and dogs were beloved by many for their unique traits. Dogs, known for their loyalty, were particularly cherished by humans. \newline \newline \textbf{Question:} Does this imply that dogs have fur?\\ \midrule
    
    DRS &  
    p: John and Mary are parents. \newline
    q: Parents are usually loving and supportive. \newline
    r: Parents are normally responsible. \newline
    s: Mary isn't loving and supportive. \newline
    t: John is not responsible. \newline
    u: Mary is responsible. \newline
    $\sim$u: Mary isn't responsible. \newline
    v: John is loving and supportive. \newline
    $\sim$v: John isn't loving and supportive. 
    & \textbf{Context:} John and Mary were expecting their first child, filled with the anticipation and excitement that all parents feel. Parents are usually loving and supportive. Parents are normally responsible. However, something seemed amiss in their relationship. Mary, usually affectionate and caring, seemed distant and uninvolved. On the other hand, John, known for his responsible nature, started neglecting his duties and became unreliable. \newline \newline \textbf{Question:} Does this imply that Mary is responsible and John is loving and supportive?\\ \midrule
    
    DRD &  
    p: Jenny and Anna are tall. \newline
    q: Tall people usually play basketball. \newline
    r: Anna is possibly an exception to this rule. \newline
    s: Jenny plays basketball. \newline
    $\sim$s: Jenny doesn't play basketball. 
    & \textbf{Context:} Jenny and Anna are known for their tall stature, which is often associated with playing basketball. However, Anna might be an exception to this norm. \newline \newline \textbf{Question:} Does this entail that Jenny plays basketball?\\ \midrule 
    
    DRO & 
    p: Hummingbirds are birds. \newline 
    q: Birds migrate south for the winter. \newline  
    r: Hummingbirds do not migrate south for the winter. \newline 
    s: All other birds than hummingbirds migrate south for the winter. \newline 
    $\sim$s: All other birds than hummingbirds don't migrate south for the winter.
    & \textbf{Context:} In the bird kingdom, there are many different species that possess unique characteristics. One such species is the hummingbird, known for its ability to hover in mid-air and its vibrant colors. While most birds engage in the annual migration south for the winter, the hummingbird chooses to stay put and brave the cold weather. This decision sets the hummingbird apart from its fellow avian companions, as it relies on its resilience and resourcefulness to survive the harsh conditions. \newline \newline \textbf{Question:} Does this mean that all other birds than hummingbirds migrate south for the winter?\\ \midrule
    
    RE1 &  
    p: Cats, dogs, and horses are animals. \newline
    q: Animals are usually considered to be intelligent creatures. \newline
    r: At least one of the cats or dogs is not considered intelligent. \newline
    s: Horses are considered to be intelligent creatures. \newline
    $\sim$s: Horses aren't considered to be intelligent creatures. \newline
    t: Exactly one of the cats or dogs is not considered intelligent. \newline
    $\sim$t: Exactly one of the cats or dogs is considered intelligent. 
    & \textbf{Context:} In a world where animals are often regarded as intelligent creatures, there is a captivating tale that revolves around cats, dogs, and horses. It is commonly believed that most animals possess a level of intellect. However, there is an intriguing twist to this belief as it is known that either cats or dogs are not considered particularly intelligent. As the story unfolds, we delve into the lives of these remarkable creatures, their interactions, and the unique qualities that each of them possesses. \newline \newline \textbf{Question:} Does this entail that horses are considered to be intelligent creatures and exactly one of the cats or dogs is not considered intelligent?\\ \midrule
    
    RE2 &  
    p: cats normally meow. \newline
    q: At least one species of cat doesn't meow. \newline
    r: Exactly one species of cat doesn't meow. \newline
    $\sim$r: Exactly one species of cat meows. 
    & \textbf{Context:} In the realm of cat communication, meowing serves as a fundamental aspect of their vocal repertoire. However, intriguingly enough, there exists a distinct species of cat that deviates from this conventional norm. This peculiar feline defies the expectations associated with its kind by refraining from emitting any meows whatsoever. \newline \newline \textbf{Question:} Does this entail that exactly one species of cat doesn't meow?\\ \midrule
    
    RE3 & 
    p: Cars have four wheels. \newline
    q: wheels normally have spokes. \newline
    r: at least one wheel does not have spokes. \newline
    s: Cars have four wheels with spokes. \newline
    $\sim$s: Cars don't have four wheels with spokes. 
    & \textbf{Context:} In a world where cars were known for having four wheels, it was considered a common fact that wheels typically came equipped with spokes. However, amidst this widespread understanding, there was an exception. At least one wheel defied this norm and stood out from the rest by not having any spokes at all. \newline \newline \textbf{Question:} Does this imply that cars have four wheels with spokes?\\ \midrule
    
    RAP & 
    p: John asserts that Sally was in the store. \newline
    q: Jane asserts that Sally was not in the store. \newline
    r: John's evidence is more reliable than Jane's. \newline
    $\sim$r: John's evidence is less reliable than Jane's. \newline
    s: Sally was in the store. \newline
    $\sim$s: Sally wasn't in the store. 
    & \textbf{Context:} In the midst of a heated argument, John adamantly claims that Sally was present at the store. However, Jane strongly opposes John's assertion, insisting that Sally was indeed absent from the store. \newline \newline \textbf{Question:} If John's evidence is more reliable than Jane's, does this mean that Sally was in the store?\\ \bottomrule

\end{tabular}
}
\vspace{0.1cm}
\caption{Illustrative examples of NL logical context and questions created using sentences that are generated in \textsection \ref{sec:sent_gen} for NM logic for BQA.}
\label{tab:NL_conversion_examples_2}
\end{table*}
\begin{table*}[htbp]
  \centering
    \resizebox{0.95\linewidth}{!}{
    \begin{tabular}{c|c|cc|cc|cc|cc|cc}\toprule
    \multicolumn{1}{c|}{\multirow{2}[0]{*}{Type}} & \multicolumn{1}{c|}{\multirow{2}[0]{*}{Rule}} & \multicolumn{2}{c|}{Llama-2} & \multicolumn{2}{c|}{Mistral} & \multicolumn{2}{c|}{Gemini} & \multicolumn{2}{c|}{ChatGPT} & \multicolumn{2}{c}{GPT-4} \\ \cmidrule{3-12}
          &  & $A(No)$ & $A(Yes)$ & $A(No)$ & $A(Yes)$ & $A(No)$ & $A(Yes)$ & $A(No)$ & $A(Yes)$ & $A(No)$ & $A(Yes)$ \\ \midrule
    \multicolumn{1}{c|}{\multirow{8}[0]{*}{PL}} & HS    & 100.0 & 25.3  & 74.3  & 16.7  & 83.3  & 26.9  & 84.2  & 27.9  & 93.8  & 53.1 \\
          & DS    & 66.7  & 26.9  & 76.6  & 66.7  & 95.2  & 32.8  & 87.5  & 33.3  & 92.2  & 93.8 \\
          & CD    & 100.0 & 55.6  & 97.1  & 42.2  & 98.1  & 73.1  & 100.0 & 57.1  & 100.0 & 80.0 \\
          & DD    & 93.8  & 79.2  & 61.1  & 59.1  & 94.7  & 90.0  & 64.3  & 83.3  & 64.3  & 83.3 \\
          & BD    & 76.9  & 25.4  & 76.4  & 37.5  & 77.8  & 24.6  & 82.4  & 30.4  & 85.1  & 76.9 \\
          & MT    & 100.0 & 25.6  & 89.3  & 32.7  & 92.9  & 28.1  & 100.0 & 30.3  & 91.5  & 48.5 \\
          & MI    & 55.6  & 51.6  & 57.9  & 57.1  & 66.7  & 66.7  & 64.0  & 73.3  & 90.5  & 94.7 \\
          & CT    & 84.7  & 52.6  & 97.4  & 45.2  & 97.4  & 48.7  & 85.7  & 36.8  & 96.7  & 90.0 \\ \midrule
          & \textbf{Avg} & \textbf{84.7} & \textbf{42.8} & \textbf{78.8} & \textbf{44.7} & \textbf{88.3} & \textbf{48.9} & \textbf{83.5} & \textbf{46.6} & \textbf{89.2} & \textbf{77.5} \\ \midrule\midrule
    \multicolumn{1}{c|}{\multirow{9}[0]{*}{FOL}} & EG    & 90.9  & 27.5  & 76.8  & 36.4  & 88.9  & 27.6  & 77.4  & 29.6  & 90.4  & 53.6 \\
          & UI    & 75.0  & 25.8  & 79.7  & 83.3  & 88.0  & 31.5  & 85.7  & 33.3  & 92.1  & 88.2 \\
          & MP    & 100.0 & 69.0  & 100.0 & 64.5  & 95.2  & 100.0 & 100.0 & 100.0 & 100.0 & 95.2 \\
          & HS    & 100.0 & 100.0 & 100.0 & 83.3  & 100.0 & 100.0 & 100.0 & 100.0 & 100.0 & 100.0 \\
          & DS    & 100.0 & 51.3  & 97.3  & 44.2  & 87.5  & 65.0  & 100.0 & 50.0  & 92.9  & 66.7 \\
          & CD    & 90.5  & 94.7  & 61.9  & 63.2  & 81.0  & 84.2  & 81.0  & 84.2  & 74.1  & 100.0 \\
          & DD    & 80.0  & 25.3  & 74.7  & 0.0   & 88.0  & 30.9  & 78.3  & 29.4  & 85.1  & 76.9 \\
          & BD    & 72.7  & 58.6  & 77.8  & 58.1  & 81.8  & 94.1  & 79.2  & 93.8  & 90.5  & 94.7 \\
          & MT    & 85.7  & 69.2  & 57.7  & 64.3  & 82.4  & 73.9  & 86.4  & 94.4  & 66.7  & 100.0 \\ \midrule
          & \textbf{Avg} & \textbf{88.3} & \textbf{57.9} & \textbf{80.7} & \textbf{55.2} & \textbf{88.1} & \textbf{67.5} & \textbf{87.5} & \textbf{68.3} & \textbf{88.0} & \textbf{86.2} \\ \midrule \midrule
    \multicolumn{1}{c|}{\multirow{8}[0]{*}{NM}} & DRI   & 86.2  & 70.6  & 62.5  & 62.5  & 80.4  & 90.9  & 74.0  & 90.0  & 95.2  & 100.0 \\
          & DRS   & 47.4  & 47.6  & 77.3  & 83.3  & 61.9  & 70.6  & 60.7  & 75.0  & 90.0  & 90.0 \\
          & DRD   & 100.0 & 66.7  & 100.0 & 62.5  & 78.6  & 68.0  & 63.6  & 55.2  & 70.4  & 92.3 \\
          & DRO   & 58.1  & 14.3  & 66.7  & 7.7   & 62.8  & 9.1   & 64.1  & 14.6  & 76.5  & 33.3 \\
          & RE1   & 60.9  & 64.7  & 58.6  & 72.7  & 66.7  & 85.7  & 60.6  & 100.0 & 87.0  & 100.0 \\
          & RE2  & 0.0   & 25.0  & 88.0  & 30.9  & 84.8  & 32.6  & 77.8  & 25.8  & 86.5  & 46.4 \\
          & RE3 & 52.6  & 52.4  & 68.0  & 80.0  & 82.4  & 71.4  & 62.1  & 81.8  & 61.5  & 71.4 \\
          & RAP   & 50.0  & 50.0  & 65.5  & 90.9  & 62.5  & 100.0 & 55.6  & 100.0 & 87.0  & 100.0 \\ \midrule
          & \textbf{Avg} & \textbf{56.9} & \textbf{48.9} & \textbf{73.3} & \textbf{61.3} & \textbf{72.5} & \textbf{66.0} & \textbf{64.8} & \textbf{67.8} & \textbf{81.8} & \textbf{79.2} \\ \bottomrule
    \end{tabular}%
    }
    \vspace{0.2cm}
    \caption{Performance of LLMs in few-shot setting in terms of label-wise accuracy on LogicBench(Eval), where $A(Yes)$ and $A(No)$ denote the accuracy for the $Yes$ and $No$ labels, respectively.}
    \label{tab:few_shot_result}%
\end{table*}%

\paragraph{LLMs sometimes overlook contextual information}

In this case study, we manually examined all reasoning chains for PL and NM generated by GPT-4. The example below illustrates the observed pattern. Specifically, in step 5 of the reasoning chain, the model deviates from the provided context and introduces unrelated information, resulting in an incorrect prediction. Although these patterns are not prevalent, we found that $\sim1\%$ of the total analyzed reasoning chains exhibited this pattern in PL and NM for BQA, and $\sim0.8\%$ for MCQA.

\begin{tcolorbox}[colback=white]
\textbf{Rule:} Modus Tollens (PL)

\textbf{Context:} James had always been forgetful, especially when it came to wearing a sweater. his mother had repeatedly told him, 'If he forgets to wear a sweater, then he will catch a cold.' throughout the day, he went about his activities, oblivious to the dropping temperature. hours later, he returned home, but James didn't catch a cold.

\textbf{Question:} Does this infer that James didn't forget to wear a sweater? 

\textbf{Gold Answer:} Yes

\textbf{Reasoning steps:}\\
1. James is known to be forgetful about wearing a sweater.\\
2. His mother stated that if he forgets to wear a sweater, he will catch a cold.\\
3. The temperature dropped, which would increase the likelihood of catching a cold if he were not wearing a sweater.\\
4. Despite the dropping temperature, James did not catch a cold.\\
5. \hlred{The absence of a cold does not necessarily mean that James remembered to wear a sweater, as there could be other reasons why he did not catch a cold (e.g., he may have a strong immune system, spent little time outdoors, or was in a warm environment).}\\
6. Therefore, the fact that James did not catch a cold does not logically infer that he remembered to wear a sweater.

Answer: No
\end{tcolorbox}

\paragraph{PL \textit{vs.} FOL} 
To get an indication of this we gave GPT-4 prompt - ``Give twenty statements that have `if' and `then' in them''. From the results, we can observe that 13 of the 20 sentences in response to the prompt were FOL kind and only 7 were propositional kind. This shows that LLMs’ comprehend simple FOL sentences, thus, showing high overall accuracy on simpler FOL compared to PL. However, we interestingly observed that, as the depth increases, PL-based examples carrying a specific case/story become easier to understand compared to FOL-based generalized context. 

\paragraph{Negations are hard to understand when embedded with logical rules.}

Here, Table \ref{tab:neg_example} presents reasoning chains corresponding to PL and FOL supporting the effect of negation on their predictions.

\begin{table*}[!htbp]
  \centering
  \resizebox{0.95\linewidth}{!}{
    \begin{tabular}{c|m{8cm}|c|m{10cm}} \toprule
    Rule  & Context/Question & Actual/Pred & Reasoning Chains \\ \midrule
    \textbf{PL(MT)} & Context: if elijah has a new job, then he will get a raise. however, he would not be receiving the raise he had anticipated.\newline{}Question: does this infer that elijah doesn't have a new job? & Yes/No & 1. The context provides a conditional statement: If Elijah has a new job, then he will get a raise.\newline{}2. The context also states that Elijah would not be receiving the raise he had anticipated.\newline{}3. The second statement does not negate the first; it only states that the raise is not as expected.\newline{}4. The lack of the anticipated raise does not necessarily mean Elijah doesn't have a new job; it could mean that the raise with the new job is different from what he expected.\newline{}5. Therefore, the information provided does not allow us to infer that Elijah doesn't have a new job based solely on the raise situation. \\ \midrule
    \textbf{PL(MT)} & Context: james had always been forgetful, especially when it came to wearing a sweater. his mother had repeatedly told him, 'if he forgets to wear a sweater, then he will catch a cold.' throughout the day, he went about his activities, oblivious to the dropping temperature. hours later, he returned home, but james didn't catch a cold.\newline{}Question: does this infer that james didn't forget to wear a sweater? & Yes/No & 1. The mother's rule establishes a conditional relationship: If James forgets to wear a sweater (condition), then he will catch a cold (result).\newline{}2. It is stated that James did not catch a cold upon returning home.\newline{}3. According to the mother's rule, if the result (catching a cold) did not occur, we cannot conclusively determine whether the condition (forgetting to wear a sweater) happened or not.\newline{}4. The absence of the result (catching a cold) does not necessarily mean the condition (forgetting to wear a sweater) did not occur, as there could be other factors at play that prevented James from catching a cold.\newline{}5. Therefore, from the information given, we cannot infer with certainty that James remembered to wear a sweater. \\ \midrule\midrule
    \textbf{FOL(DD)} & Context: if an individual operates a vehicle with caution, they will prevent any accidents from occurring. however, if they decide to accelerate and exceed the speed limit, they will be issued a fine. there is a certainty that at least one of the following statements is true: either john will not be able to avoid an accident or he will not face a penalty. it is unclear which of these statements is accurate, as it could be only one of them, or it is possible that both of them hold true.\newline{}Question: can we say at least one of the following must always be true? (a) john doesn't drive carefully and (b) he doesn't speed up & Yes/No & 1. According to the context, if an individual operates a vehicle with caution, they will prevent any accidents from occurring.\newline{}2. Also, if they decide to accelerate and exceed the speed limit, they will be issued a fine.\newline{}3. It is stated that there is a certainty that at least one of the following statements is true: either John will not be able to avoid an accident or he will not face a penalty.\newline{}4. It is unclear which of these statements is accurate, as it could be only one of them, or it is possible that both of them hold true.\newline{}5. The question asks if we can say at least one of the following must always be true: (a) John doesn't drive carefully and (b) he doesn't speed up.\newline{}6. Based on the given information, we cannot say that at least one of the statements (a) and (b) must always be true. \\ \midrule
    \textbf{FOL(DD)} & Context: if someone is taking care of thier health, then they will be fit. however, indulging in unhealthy habits can make individuals susceptible to various diseases. the truth is, we can be certain about at least one of the following possibilities: either jenny won't be fit and healthy, or she won't be prone to diseases. it is important to note that we are unaware of which statement is accurate. it could be the case that only the first statement is true, only the second statement is true, or both statements hold validity.\newline{}Question: can we say at least one of the following must always be true? (a) jenny takes care of her health and (b) she doesn't indulge in unhealthy habits & No/Yes & 1. According to the context, if someone takes care of their health, they will be fit.\newline{}2. Indulging in unhealthy habits can make individuals susceptible to various diseases.\newline{}3. The context states that we can be certain about at least one of the following possibilities: either Jenny won't be fit and healthy or she won't be prone to diseases.\newline{}4. It is important to note that we are unaware of which statement is accurate.\newline{}5. The possibilities are:\newline{}   a) Jenny takes care of her health and is fit.\newline{}   b) Jenny doesn't indulge in unhealthy habits and is not prone to diseases.\newline{}   c) Both statements hold validity.\newline{}6. Therefore, at least one of the following must always be true: (a) Jenny takes care of her health and (b) she doesn't indulge in unhealthy habits. \\ \bottomrule
    \end{tabular}%
    }
    \caption{Case study of reasoning chains for PL(MT) generated from ChatGPT, and FOL(DD) generated from GPT-4.}
  \label{tab:neg_example}%
\end{table*}%

\paragraph{Performance of BQA \textit{vs.} MCQA} 
By analyzing the reasoning chains of CT and DD for PL, we observe that LLMs often employ the elimination method in MCQA. They prioritize options to select the most appropriate conclusion, thereby improving performance. Conversely, for NM, the models typically struggle to differentiate between two options and tend to hallucinate by assuming information, resulting in lower performance. We present reasoning chains corresponding to PL and NM supporting their behavior on both BQA and MCQA tasks at \url{https://github.com/Mihir3009/LogicBench}.

\paragraph{Results on Yi-34B}
\begin{table*}[!htbp]
\centering
\resizebox{0.5\linewidth}{!}{
\begin{tabular}{c|c|cc|c}
\toprule
\textbf{Logic} &\multirow{2}{*}{\textbf{Rule}} &\multicolumn{2}{|c|}{\textbf{BQA}} &\textbf{MCQA} \\ \cmidrule{3-5}
 \textbf{Type} & & \textbf{A(No)} & \textbf{A(Yes)} & \textbf{Acc.} \\
\midrule
 \multirow{8}{*}{PL} & HS & $98.81_{0.01}$ & $92.13_{0.02}$ & $96.67_{0.03}$ \\
 & DS & $63.61_{0.06}$ & $87.25_{0.07}$ & $90.00_{0.05}$ \\
 & CD & $88.88_{0.02}$ & $41.04_{0.05}$ & $86.67_{0.03}$ \\
 & DD & $73.69_{0.01}$ & $19.53_{0.01}$ & $63.33_{0.08}$ \\
 & BD & $83.62_{0.03}$ & $43.38_{0.04}$ & $65.00_{0.05}$ \\
 & MT & $46.88_{0.03}$ & $31.55_{0.05}$ & $58.33_{0.06}$ \\
 & MI & $84.49_{0.01}$ & $31.57_{0.01}$ & $71.67_{0.06}$ \\
 & CT & $95.48_{0.04}$ & $59.15_{0.01}$ & $70.00_{0.13}$ \\ \midrule
 & \textbf{Avg} & \textbf{$79.43_{0.03}$} & \textbf{$50.7_{0.03}$} & \textbf{$75.21_{0.06}$} \\ \midrule\midrule
 
 \multirow{10}{*}{FOL} & EG & ~$100.0_{0.0}$ & $98.33_{0.03}$ & $88.33_{0.03}$ \\
 & UI & $90.32_{0.00}$ & $94.64_{0.00}$ & $88.33_{0.08}$ \\
 & MP & $84.87_{0.02}$ & ~$100.0_{0.0}$ & $93.33_{0.08}$ \\
 & HS & $96.67_{0.02}$ & $94.52_{0.00}$ & $76.67_{0.03}$ \\
 & DS & $77.42_{0.03}$ & $90.28_{0.09}$ & $55.00_{0.05}$ \\
 & CD & $90.41_{0.01}$ & $40.39_{0.02}$ & $81.67_{0.08}$ \\
 & DD & $72.77_{0.02}$ & $22.20_{0.03}$ & $50.00_{0.20}$ \\
 & BD & $79.60_{0.03}$ & $30.05_{0.06}$ & $60.00_{0.05}$ \\
 & MT & $50.95_{0.02}$ & $56.11_{0.21}$ & $68.33_{0.19}$ \\ \midrule
 & \textbf{Avg} & \textbf{$82.56_{0.02}$} & \textbf{$69.61_{0.05}$} & \textbf{$73.52_{0.09}$} \\ \midrule \midrule
 
 \multirow{9}{*}{NM} & DRI & $86.10_{0.08}$ & $98.25_{0.03}$ & $60.00_{0.09}$ \\
 & DRS & $73.79_{0.00}$ & $10.37_{0.10}$ & $30.00_{0.05}$ \\
 & DRD & $85.76_{0.05}$ & ~$100.0_{0.0}$ & $55.00_{0.09}$ \\
 & DRO & $70.37_{0.04}$ & ~$100.0_{0.0}$ & $33.33_{0.06}$ \\
 & REI & $83.80_{0.03}$ & $44.31_{0.09}$ & $73.33_{0.08}$ \\
 & REII & $63.27_{0.02}$ & $76.03_{0.06}$ & $63.33_{0.08}$ \\
 & REIII & $63.89_{0.03}$ & $88.33_{0.13}$ & $56.67_{0.03}$ \\
 & RAP & $68.15_{0.06}$ & $87.12_{0.03}$ & $38.33_{0.08}$ \\ \midrule
 & \textbf{Avg} & \textbf{$74.39_{0.04}$} & \textbf{$75.55_{0.05}$} & \textbf{$51.25_{0.07}$} \\
\bottomrule
\end{tabular}
}
\caption{Evaluation of Yi-34B in terms of accuracy on \textit{LogicBench(Eval)}$_{BQA}$ and \textit{LogicBench(Eval)}$_{MCQA}$.}
\label{tab:yi_results}
\end{table*}
Table \ref{tab:yi_results} provides results for both BQA and MCQA tasks on \textit{LogicBench}.

\end{document}